\documentclass{fairmeta} %

\usepackage{latexsym}
\usepackage[utf8]{inputenc}
\usepackage{etoc}
\usepackage{wrapfig}
\usepackage{amsmath,amssymb}
\usepackage{booktabs}
\usepackage{multirow}
\usepackage{colortbl}
\usepackage{enumitem}
\usepackage{tcolorbox}
\usepackage{adjustbox}
\usepackage{placeins}  %
\setlist{nosep,leftmargin=*}

\newif\ifmarkchanges \markchangesfalse
\newcommand{\added}[1]{\ifmarkchanges\textcolor{red}{#1}\else#1\fi}
\newcommand{\moved}[1]{\ifmarkchanges\textcolor{blue}{#1}\else#1\fi}

\newcommand{\method}{PremiseCheck}  %
\newcommand{\methodhd}{\begin{tabular}[c]{@{}c@{}}Premise-\\Check\end{tabular}}  %
\newcommand{\jev}{Jev}
\newcommand{\kev}{Kev-27B}
\newcommand{\gpt}{GPT-5.5}
\newcommand{\mini}{GPT-5.4-mini}
\newcommand{\rc}{ReviewCritique}
\newcommand{\rs}{ReviewScore}
\newcommand{\prb}{PeerReview Bench}
\newcommand{\bench}{ReviewFact}  %

\definecolor{rowgray}{gray}{0.45}
\newcommand{\dimrow}[1]{\textcolor{rowgray}{#1}}
\definecolor{ourrow}{RGB}{232,240,252}
\definecolor{errcell}{RGB}{251,234,230}
\definecolor{okcell}{RGB}{230,239,246}
\definecolor{effcell}{RGB}{226,160,90}
\definecolor{deltapos}{RGB}{0,128,60}
\definecolor{deltaneg}{RGB}{192,0,0}
\newcommand{\up}[1]{\textsubscript{\textcolor{deltapos}{$+$#1}}}
\newcommand{\down}[1]{\textsubscript{\textcolor{deltaneg}{$-$#1}}}

\newtcolorbox{takeaway}{colback=ourrow,colframe=ourrow!60!black,boxrule=0.4pt,arc=2pt,
  left=4pt,right=4pt,top=2pt,bottom=2pt}

\usepackage{listings}
\tcbuselibrary{listings,breakable,skins}
\definecolor{llmframe}{RGB}{168,112,42}
\definecolor{llmback}{RGB}{252,243,227}
\newtcblisting{promptbox}[1]{listing only,breakable,enhanced,
  colback=llmback!45,colframe=llmframe,boxrule=0.5pt,arc=2pt,
  left=3pt,right=3pt,top=2pt,bottom=2pt,before skip=5pt,after skip=6pt,
  title={\scriptsize\sffamily\bfseries #1},fonttitle=\color{white},
  coltitle=white,colbacktitle=llmframe,toptitle=1pt,bottomtitle=1pt,
  listing options={basicstyle=\ttfamily\scriptsize\linespread{0.95}\selectfont,breaklines=true,
    columns=fullflexible,breakindent=1em,aboveskip=0pt,belowskip=0pt}}

\renewcommand{\topfraction}{0.85}\renewcommand{\textfraction}{0.12}\renewcommand{\floatpagefraction}{0.75}
\newenvironment{nolinenumbers}{}{}  %

\title{\method{}: Efficient Fact-Checking of Human and AI Peer Reviews}

\author[1]{Dingjie Song}
\author[1]{Zhiling Yan}
\author[1]{Ziqian Zhang}
\author[2,3]{Yuxuan Zhang}
\author[1,\dagger]{Lichao Sun}

\affiliation[1]{Lehigh University}
\affiliation[2]{University of British Columbia}
\affiliation[3]{Vector Institute}

\contribution[\dagger]{Corresponding author}

\abstract{Peer review is under strain from the growth of submissions, and review comments written by humans or AI can misstate what a paper says.
We introduce \bench{}, a benchmark that labels such factual errors in review comments under one definition across NLP, machine learning, and the natural sciences, and that measures cost and latency alongside accuracy.
A frontier LLM checks comments well but must read the whole paper for each check, while cheap decision models, which score whether a statement holds in a text, fall far short on whole comments that mix several implicit assertions.
\method{} is a framework that combines the two.
An LLM rewrites each comment as explicit premises without reading the paper, and a decision model checks each premise against the paper.
When each comment is checked on its own, \method{} reaches accuracy comparable to \gpt{} across all three fields at about half its cost, and on NLP reviews at a quarter of its cost and less than half its latency.
}

\metadata[Code]{\url{https://github.com/OpenLAIR/PremiseCheck}}
\metadata[Data]{\url{https://huggingface.co/datasets/songdj/ReviewFact}}
\metadata[Website]{\url{https://openlair.github.io/premisecheck/}}

\hypersetup{
  pdftitle={PremiseCheck: Efficient Fact-Checking of Human and AI Peer Reviews},
  pdfauthor={Dingjie Song, Zhiling Yan, Ziqian Zhang, Yuxuan Zhang, Lichao Sun}
}

\newcommand{\onelogo}[1]{\raisebox{-0.5\height}{\includegraphics[height=0.55cm]{logos/#1}}}
\renewcommand{\titlelogos}{\onelogo{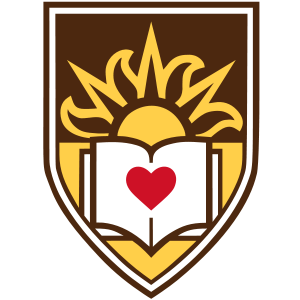}\;\onelogo{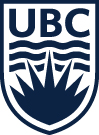}\;\onelogo{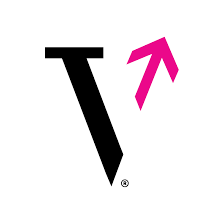}}

\begin{document}
\maketitle

\etocdepthtag.toc{mtmain}
\section{Introduction}
\label{sec:intro}

\begin{wrapfigure}[24]{r}{0.44\textwidth}
\centering
\vspace{-\baselineskip}\setlength{\abovecaptionskip}{3pt}
\includegraphics[width=\linewidth]{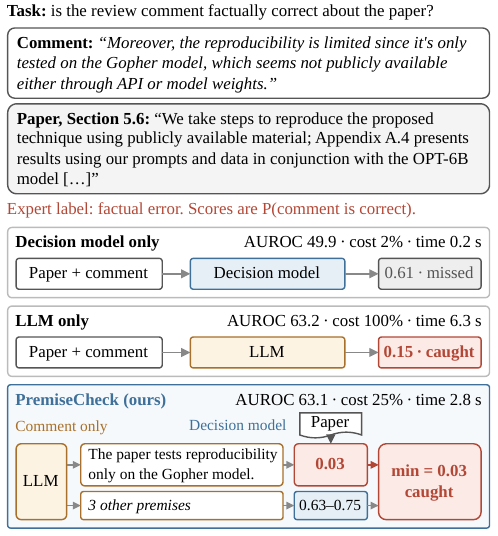}
\caption{Three ways to check a review comment, shown on a real comment from \bench{}. AUROC, cost, and time are measured on the NLP reviews of \bench{}, with cost relative to the LLM-only row, which is the reference in our experiments. The LLM is \gpt{} and the decision model is \jev{}.}
\label{fig:paradigms}
\end{wrapfigure}

The number of research papers has grown explosively in recent years, and reviewing them now places a heavy burden on the scientific community \citep{shah2022challenges,kuznetsov2024nlp,kim2025reviewcrisis}.
A review comment is useful only if what it says about the paper is true, yet factual errors appear regularly in both human-written and AI-written comments \citep{du2024reviewcritique,ryu2025reviewscore,kim2026peerreviewbench}, and AI systems now help write a growing share of reviews \citep{liang2024monitoring,thakkar2025reviewfeedback,biswas2026aaai26}.
Such errors mislead authors and distort decisions, and at the present scale they can no longer be caught by hand.
\added{An area chair cannot reread each paper to check every comment, and an author can correct an error only in the rebuttal.}

We therefore study the fact-checking of review comments as a task of its own.
Given a paper and one comment from a review of it, a system scores how likely the comment says something that the paper does not support.
Errors of judgment, such as whether a concern matters, are out of scope.
Existing expert annotations of review comments label factual errors inconsistently, if at all.
Our benchmark, \bench{}, builds on them \citep{du2024reviewcritique,ryu2025reviewscore,kim2026peerreviewbench} and labels every comment under one definition of factual error.
Because a system would run on every comment of every review, \bench{} evaluates cost and latency together with accuracy.
The task is hard because each error is something an expert reviewer overlooked or misread, and the evidence against it may lie anywhere in a long paper.
\added{A comment also rarely states its claim directly, as a question such as ``Did the authors test other models?'' presupposes that the paper does not test them.}

When each comment is checked on its own, existing systems are either accurate or cheap, but not both.
A frontier LLM that judges each comment on its own \citep{zheng2023judging,ryu2025reviewscore,thakkar2025reviewfeedback} must reread the full paper for every comment, a cost that multiplies across tens of thousands of reviews.
Instead of generating text, a decision model such as \jev{} returns the probability that a statement holds in a document, typically within a fraction of a second and at negligible cost \citep{typesafe2026blog,typesafe2026docs}.
Yet on whole comments, \jev{} performs at chance level on the NLP reviews and well below an LLM on the other two parts.
A review comment mixes several assertions, many implicit, whereas a decision model, like single-claim fact-checkers \citep{laban2022summac,zha2023alignscore,tang2024minicheck}, is reliable only on explicit, self-contained claims.
\added{On single premises that human annotators wrote out from review comments, however, \jev{} is as accurate as \gpt{} (Appendix~\ref{app:capmap}).
Our key insight is that understanding what a comment asserts needs an LLM but not the paper, whereas reading the paper is the expensive part and a decision model can do it once the assertions are explicit.}

\method{} is a framework that combines the two kinds of models.
An LLM rewrites each comment as a few explicit premises without reading the paper \citep{min2023factscore,wei2024safe,ryu2025reviewscore}.
A decision model then decides how to check each premise and verifies it against the paper, and the comment receives the score of its weakest premise.
Figure~\ref{fig:paradigms} contrasts \method{} with either kind of model alone.
\added{With a medium-reasoning decomposer, \method{} stays within 0.3 AUROC points of holistic \gpt{} or above it on every part of \bench{} at about half its cost, and on the NLP reviews a decomposer without reasoning cuts the cost to a quarter while remaining non-inferior.
The accuracy comes from decomposition rather than from one model, since four decision models from different providers, including the open-weight \kev{}, come within one point of holistic \gpt{} as verifiers.}

Our contributions are as follows.
\begin{itemize}
  \item We propose \method{}, a framework that turns the verification of a review comment into decision problems, combining the flexibility of an LLM with the efficiency of a decision model.
  \added{A decision model checks each premise against the paper or against general knowledge and asks several phrasings of its question in one call.}
  \item We define the task of fact-checking single review comments and build \bench{}, a benchmark that labels factual errors in human-written and AI-written review comments across NLP, machine learning, and the natural sciences and measures cost and latency alongside accuracy.
  \added{\bench{} holds 9,267 comments on 217 papers.}
  \item We validate \method{} on all three parts of \bench{}, where it checks each comment on its own with accuracy comparable to \gpt{} at a fraction of its cost.
  \added{A cascade costs more than \method{} yet is less accurate on every part.}
\end{itemize}

\section{Related Work}
\label{sec:related}

\subsection{Factual Errors in Review Comments}
As LLMs write reviews and critique human ones \citep{liang2024feedback,darcy2024marg,zhu2025deepreview,thakkar2025reviewfeedback,zhuang2025survey}, checking what a review says about a paper matters more.
Unlike benchmarks of errors in papers \citep{son2025spot,zhang2025critical}, \rc{} \citep{du2024reviewcritique}, \rs{} \citep{ryu2025reviewscore}, \prb{} \citep{kim2026peerreviewbench}, and ClaimCheck \citep{ou2025claimcheck} annotate review comments against the paper.
These datasets serve broader goals and label factual errors inconsistently, if at all, whereas \bench{} applies one definition across three fields and also scores cost and latency.
\added{Systems that generate reviews aim to write better comments, whereas \method{} checks comments that have already been written.}

\subsection{Claim Decomposition and Grounded Checking}
Fact verification commonly splits a text into atomic claims and checks each one against a source \citep{min2023factscore,wei2024safe}, often with a small trained fact-checking model \citep{tang2024minicheck}.
\rs{} and Peerispect \citep{ghorbanpour2026peerispect} bring this recipe to peer review with an LLM as the verifier.
\added{When an LLM reads the paper to verify each premise, decomposition multiplies the number of expensive calls instead of reducing it.}
\method{} instead has an LLM write premises without reading the paper and a decision model check them.
Unlike a trained fact-checking model, a decision model such as \jev{} \citep{typesafe2026blog,typesafe2026docs} takes a question as input, so \method{} can route each premise and phrase its check in several ways.

\subsection{Cost-Aware Inference}
Cascades and routers send each query to a cheap or an expensive LLM \citep{chen2023frugalgpt,ong2025routellm,dekoninck2025cascaderouting,jung2025trustorescalate}, and compound AI systems and agents are now optimized and evaluated for cost \citep{zaharia2024compound,khattab2023dspy,kapoor2024agentsmatter,erol2025costofpass}.
Test-time scaling instead spends more output tokens and time to gain accuracy \citep{wang2023selfconsistency,snell2025scaling}.
Unlike a cascade, which still sends most comments and their papers to an expensive LLM, \method{} gives the paper only to the decision model.
\added{\method{} also runs counter to test-time scaling, since its decomposer needs no reasoning.}

\begin{figure}[t]
\centering
\includegraphics[width=\textwidth]{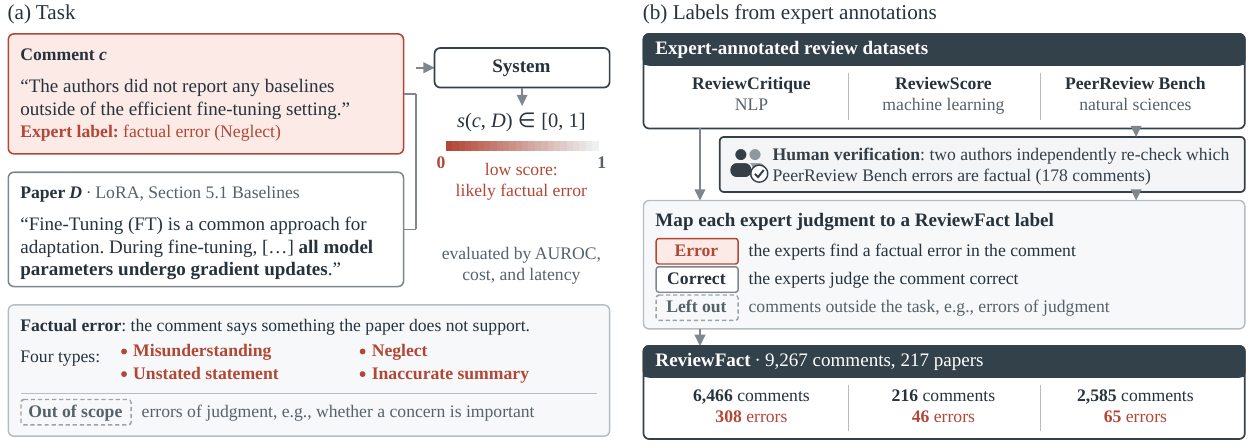}
\caption{\added{(a) The task on a real \bench{} comment that claims a baseline is missing although the paper reports it. (b) \bench{} labels from the expert annotations of three datasets. For \prb{}, two authors independently re-check which errors are factual on 178 comments and agree with the \gpt{} labels with $\kappa$ of 0.76 and 0.77.}}
\label{fig:bench-overview}
\end{figure}

\section{The \bench{} Benchmark}
\label{sec:benchmark}

\subsection{Task}
\label{sec:bench-task}

Given a paper $D$ and one comment $c$ from a review of $D$, a system outputs a score $s(c, D) \in [0,1]$, written $s(c)$ for short, where a low score means that $c$ likely contains a factual error about $D$.
A comment contains a factual error when it says something about the paper that the paper does not support.
We adopt the set $\mathcal{T}$ of four types of factual error of \rc{} \citep{du2024reviewcritique}, in which the comment misreads the paper (Misunderstanding), claims that the paper lacks something it contains (Neglect), asserts something with no basis in the paper (Unstated statement), or misstates its content (Inaccurate summary).
Errors of judgment, such as whether a concern is important, are outside the task.
\added{Figure~\ref{fig:bench-overview}a shows the task on a real comment.}

A dataset for the task is a set of labeled comments $\mathcal{D} = \{(D_j, c_j, y_j)\}_{j=1}^{N}$, where $y_j = 1$ if comment $c_j$ contains a factual error about paper $D_j$ and $y_j = 0$ otherwise.
When the source annotations provide it, an error also has a type $t_j \in \mathcal{T}$.
Systems are evaluated by AUROC with factual errors as the class to detect,
\begin{equation}
\mathrm{AUROC} = \Pr\big[s(c^{+}) < s(c^{-})\big],
\label{eq:auroc}
\end{equation}
where $c^{+}$ is a random comment with $y = 1$, $c^{-}$ a random comment with $y = 0$, and ties count one half.
We report AUROC multiplied by 100.
Because a system runs on every comment, we also measure its cost $C(c)$ in US dollars and its latency $T(c)$ in seconds per comment, and we report their means relative to the reference system of \S\ref{sec:setup}.

\subsection{Labels from Expert Annotations}
\label{sec:bench-data}

\bench{} adds no new comments or papers.
It reuses expert judgments of review comments from one dataset per field, \rc{} for NLP, \rs{} for machine learning, and \prb{} for the natural sciences \citep{du2024reviewcritique,ryu2025reviewscore,kim2026peerreviewbench}.
Human reviewers wrote comments in all three parts, which we call the NLP, machine learning, and natural-science reviews, and AI reviewers also wrote comments in the NLP and natural-science reviews.
Each dataset was built for a broader goal, so we map its annotations to our label and leave out comments outside the task, following the rules in Appendix~\ref{app:bench-annot}.
\added{Figure~\ref{fig:bench-overview}b outlines this process.}
Figure~\ref{fig:pipeline} shows a labeled sample.

\paragraph{Human verification.}
Many comments that the \prb{} experts judged incorrect are errors of judgment rather than factual errors, so \gpt{} decides from the expert's rationale whether the error is factual.
Two of the authors independently repeat this decision on the 178 comments that have a rationale, without seeing the \gpt{} labels or each other's labels (Appendix~\ref{app:bench-verify}).
The two annotators agree with a Cohen's $\kappa$ of 0.70, and \gpt{} agrees with them with $\kappa$ of 0.76 and 0.77, so we keep the \gpt{} labels.
With the labels of the annotators instead, the difference between \method{} and holistic \gpt{} on these reviews ranges from $-3.7$ to $+2.7$ and is never significant, so the \gpt{} labels do not favor the \gpt{} reference.

\subsection{Statistics}
\label{sec:bench-stats}

\noindent\begin{minipage}[t]{0.5\textwidth}
\vspace{0pt}
Figure~\ref{fig:bench-stats} summarizes the 9,267 comments on 217 papers.
Factual errors are rare, 4.8\% of the NLP comments and 2.5\% of the natural-science comments, so a system spends most of its calls on correct comments and its cost matters.
The machine learning reviews have a higher share of errors because they keep only arguments, that is, comments that give reasons for a conclusion.
\rs{} rates the premises of arguments only, so we leave out its claims and questions (Appendix~\ref{app:bench-annot}).
\moved{The median comment has 17, 48, and 70 words, and the median paper has 8.5k, 9.1k, and 14.4k tokens in the NLP, machine learning, and natural-science reviews.}
\added{A comment is thus short compared with the paper it refers to, so a system that reads the paper for every comment spends almost all of its input on the paper.}
\end{minipage}\hfill
\begin{minipage}[t]{0.47\textwidth}
\vspace{0pt}
\centering
\includegraphics[width=\linewidth]{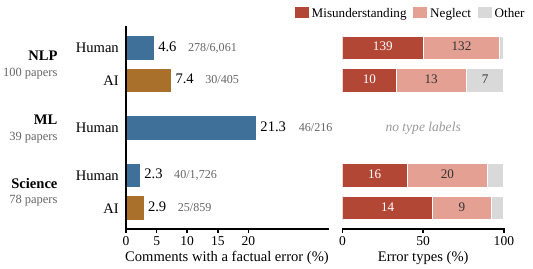}
\captionsetup{type=figure}
\caption{\bench{} by part and reviewer type. The left panel shows the share of comments with a factual error, next to the counts of errors and comments. The right panel shows the types of these errors, where Other is Unstated statement or Inaccurate summary.}
\label{fig:bench-stats}
\end{minipage}

\section{Method}
\label{sec:method}

\begin{figure*}[t]
\centering
\includegraphics[width=\textwidth]{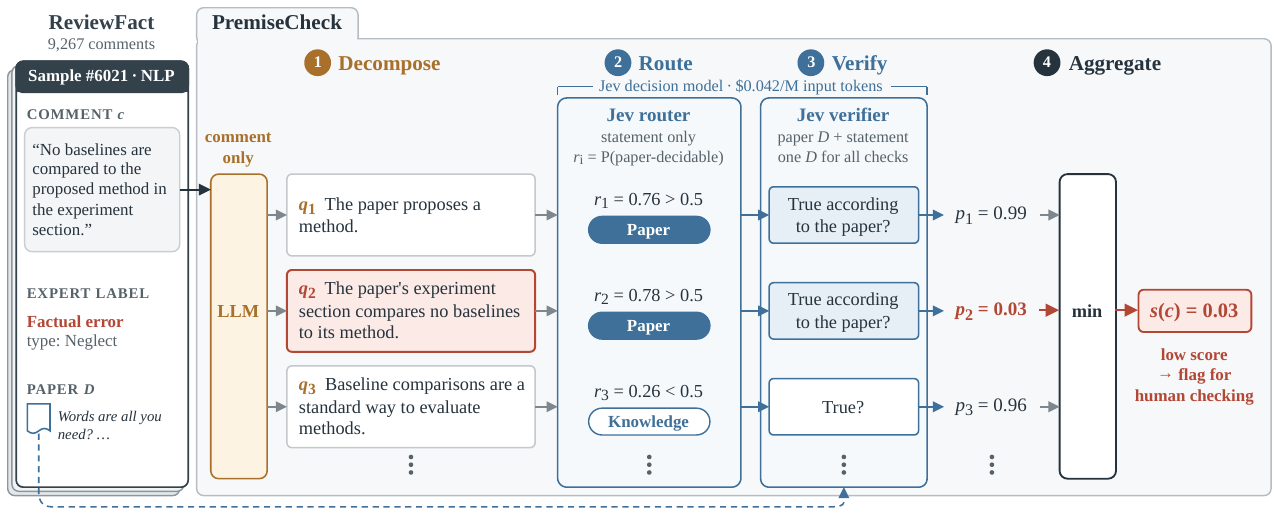}
\caption{\method{} on a \bench{} sample that experts labeled as a factual error, with premises shortened for display. Only the verifier reads the paper, and the lowest premise score, shown in red, is the comment score.}
\label{fig:pipeline}
\end{figure*}

\paragraph{Overview.}
\method{} is a framework that scores a comment $c$ on a paper $D$ for the task of \S\ref{sec:benchmark} in four steps (Figure~\ref{fig:pipeline}).
An LLM decomposes the comment into premises.
A decision model routes each premise and then verifies it.
The premise scores are aggregated into one comment score.
Only the decomposer is an LLM, and it never sees the paper.
A decision model $f$ maps an input text $x$ and a question $a$, phrased as a declarative sentence, to the probability $f(x, a) \in [0,1]$ that $a$ holds for $x$.
Any LLM and any decision model can fill the two roles, and \S\ref{sec:setup} gives the models we use.

\paragraph{Decompose.}
The LLM reconstructs the comment as an argument with a conclusion and premises
$\{q_1, \dots, q_k\} = \mathrm{LLM}(c)$, where $1 \le k \le 6$.
Each premise is one self-contained declarative sentence that names the method, table, or section it refers to.
The prompt asks for both explicit and implicit premises, including what a question presupposes about the paper, and it asks the model not to add facts that the reviewer did not assert or imply (Appendix~\ref{app:prompts}).
The input is only the comment, which averages 24 tokens on the NLP reviews, and the decomposer prompt averages about 256 tokens.

\paragraph{Route.}
Some premises are about the paper, such as ``the paper does not report results on dataset X.''
Others rely on general knowledge, such as ``method Y is a standard baseline for this task.''
Asking whether such a knowledge premise is true ``according to the paper'' would wrongly push its probability down.
The decision model therefore first estimates the probability $r_i = f(q_i, a_{\text{route}})$ that premise $q_i$ can be decided from the paper alone, where $a_{\text{route}}$ is the routing question of Appendix~\ref{app:statements}.
This routing call receives only the premise, without the paper.
With $g_i = \mathbf{1}[r_i \ge 0.5]$, premises with $g_i = 1$ are treated as paper-grounded, and the rest as knowledge premises.

\paragraph{Verify.}
The decision model then reads the paper together with the premise and answers the question that matches its route,
\begin{equation}
p_i =
\begin{cases}
f\big((D, q_i),\, a_{\text{paper}}\big) & \text{if } g_i = 1, \\
f\big((D, q_i),\, a_{\text{know}}\big) & \text{if } g_i = 0,
\end{cases}
\label{eq:verify}
\end{equation}
where $a_{\text{paper}}$ is ``The proposition is true according to the paper,'' $a_{\text{know}}$ is ``The proposition is true,'' and the proposition is $q_i$.
On the natural-science reviews, the decomposer marks which premises to verify in place of routing, and a contradiction question replaces $a_{\text{paper}}$ (Appendices~\ref{app:prompts} and~\ref{app:statements}).

\paragraph{Aggregate.}
A comment that rests on a false premise contains a factual error, so the comment score is the lowest premise score,
\begin{equation}
s(c) = \min_{1 \le i \le k} \, p_i .
\label{eq:min}
\end{equation}
If the decomposer returns no premise, which happens for fewer than 1\% of the comments in the NLP reviews, we fall back to the holistic judgment of the decision model.

\paragraph{Why it is cheap.}
The cost of \method{} for one comment is that of one LLM call on the comment plus a routing call and a verification call per premise,
\begin{equation}
C(c) = C_{\text{LLM}}(c) + \sum_{i=1}^{k} \big[ C_f(q_i) + C_f(D, q_i) \big].
\label{eq:cost}
\end{equation}
The LLM call is short because it excludes the paper.
A decision model generates no text, so each $C_f$ grows only with the length of its input, and asking several questions about the same input adds almost no cost.
Each verification call reads the paper again, yet in our implementation all decision-model calls together cost \$0.0015 per comment on the NLP reviews (Figure~\ref{fig:cost}).
In holistic judgment, by contrast, the LLM reads the paper and reasons about it for every comment, so its cost $C_{\text{LLM}}(D, c)$ grows with the length of the paper at LLM prices.

\section{Experimental Setup}
\label{sec:setup}

\begin{table*}[t]
\centering
\footnotesize
\setlength{\tabcolsep}{4pt}
\newcommand{\eff}[2]{\cellcolor{effcell!#1}#2}
\begin{tabular}{@{}llccccccccc@{}}
\toprule
 & & \multicolumn{3}{c}{NLP} & \multicolumn{3}{c}{Machine learning} & \multicolumn{3}{c}{Natural science} \\
\cmidrule(lr){3-5}\cmidrule(lr){6-8}\cmidrule(l){9-11}
 & & & \multicolumn{2}{c}{Efficiency (\%)} & & \multicolumn{2}{c}{Efficiency (\%)} & & \multicolumn{2}{c}{Efficiency (\%)} \\
\cmidrule(lr){4-5}\cmidrule(lr){7-8}\cmidrule(l){10-11}
System & Reasoning & AUROC & Cost & Time & AUROC & Cost & Time & AUROC & Cost & Time \\
\midrule
\multicolumn{11}{@{}l}{\emph{Holistic judgment}} \\
\gpt{} (ref.) & medium & 63.2 & \eff{40}{100.0} & \eff{40}{100.0} & 66.9 & \eff{40}{100.0} & \eff{40}{100.0} & \textbf{67.8} & \eff{40}{100.0} & \eff{40}{100.0} \\
\dimrow{\gpt{}} & \dimrow{low} & \dimrow{61.2}\down{2.1} & \eff{30}{\dimrow{76.1}} & \eff{27}{\dimrow{67.6}} & \dimrow{64.3}\down{2.6} & \eff{40}{\dimrow{120.5}} & \eff{27}{\dimrow{67.9}} & \dimrow{67.4}\down{0.4} & \eff{30}{\dimrow{75.5}} & \eff{18}{\dimrow{46.0}} \\
\dimrow{\gpt{}} & \dimrow{none} & \dimrow{49.3}\down{14.0} & \eff{20}{\dimrow{49.9}} & \eff{11}{\dimrow{28.4}} & \dimrow{66.5}\down{0.4} & \eff{38}{\dimrow{95.7}} & \eff{10}{\dimrow{26.1}} & \dimrow{63.7}\down{4.1} & \eff{22}{\dimrow{54.2}} & \eff{5}{\dimrow{13.7}} \\
\dimrow{\mini{}} & \dimrow{low} & \dimrow{53.7}\down{9.5} & \eff{4}{\dimrow{9.8}} & \eff{12}{\dimrow{31.1}} & \dimrow{58.7}\down{8.2} & \eff{5}{\dimrow{13.5}} & \eff{15}{\dimrow{36.6}} & \dimrow{61.4}\down{6.4} & \eff{6}{\dimrow{13.8}} & \eff{14}{\dimrow{34.4}} \\
\dimrow{\jev{}} & \dimrow{--} & \dimrow{49.9}\down{13.4} & \eff{1}{\dimrow{2.1}} & \eff{1}{\dimrow{3.6}} & \dimrow{61.2}\down{5.7} & \eff{1}{\dimrow{1.6}} & \eff{1}{\dimrow{3.6}} & \dimrow{58.7}\down{9.1} & \eff{1}{\dimrow{1.6}} & \eff{1}{\dimrow{1.9}} \\
\midrule
\multicolumn{11}{@{}l}{\emph{Cascade}} \\
\dimrow{\jev{}$\rightarrow$\gpt{}} & \dimrow{medium} & \dimrow{59.1}\down{4.2} & \eff{30}{\dimrow{74.0}} & -- & \dimrow{65.7}\down{1.1} & \eff{31}{\dimrow{76.4}} & -- & \dimrow{66.7}\down{1.1} & \eff{28}{\dimrow{69.2}} & -- \\
\midrule
\multicolumn{11}{@{}l}{\emph{\method{} (decomposer: \gpt{}, verifier: \jev{})}} \\
\rowcolor{ourrow} \method{} & medium & 63.0$^\dagger$\down{0.3} & \eff{20}{51.2} & \eff{33}{81.6} & \textbf{68.7}\up{1.8} & \eff{19}{46.6} & \eff{38}{95.7} & 67.5\down{0.3} & \eff{19}{47.4} & \eff{24}{60.6} \\
\rowcolor{ourrow} \method{} & low & \textbf{64.1}$^\dagger$\up{0.8} & \eff{14}{34.3} & \eff{24}{60.1} & 63.9\down{3.0} & \eff{12}{29.2} & \eff{21}{52.9} & 65.7\down{2.1} & \eff{10}{26.2} & \eff{11}{28.1} \\
\rowcolor{ourrow} \method{} & none & 63.1$^\dagger$\down{0.2} & \eff{10}{24.6} & \eff{18}{45.0} & 63.7\down{3.2} & \eff{8}{19.5} & \eff{15}{37.5} & 66.0\down{1.7} & \eff{9}{22.3} & \eff{9}{21.9} \\
\bottomrule
\end{tabular}
\caption{Results on \bench{} with each comment checked on its own. Subscripts give the paired AUROC difference to the reference, computed before rounding, $^\dagger$ marks non-inferiority, and cost and time are relative to the reference. Reasoning refers to the \gpt{} or \mini{} call, which for \method{} is the decomposer.}
\label{tab:main}
\end{table*}

\paragraph{Models.}
Our implementation of \method{} uses \gpt{} as the decomposer and \jev{} as the decision model, which reads the paper truncated to 24k tokens.\footnote{We use \texttt{gpt-5.5-2026-04-23}, \texttt{gpt-5.4-mini-2026-03-17}, and \texttt{jev-1.13-20260917}. The LLMs run with their default sampling and give each probability as an integer from 0 to 100, and \jev{} is billed only for input tokens, at \$0.042 per million.}
On the machine learning and natural-science reviews, the decomposer uses the prompt variants of Appendix~\ref{app:prompts}.
Our main setting scores each comment on its own, as an assistant would check a comment as soon as the reviewer writes it, before the rest of the review exists.
Checking all comments of a finished review together is a different setting, which we evaluate in \S\ref{sec:batch}.
Appendix~\ref{app:setup} gives the prices of the LLMs.

\paragraph{Baselines.}
\emph{Holistic judgment} gives a model the paper and the comment and asks for the probability that the comment's main criticism is factually correct.
We run it with \gpt{} at medium, low, and no reasoning, with \mini{}, and with \jev{}.
The reference system is holistic \gpt{} at medium reasoning.
A \emph{cascade} scores all comments with holistic \jev{} and sends the 70\% it finds most suspicious to holistic \gpt{}.

\paragraph{Metrics.}
We call a system non-inferior to the reference when the lower bound of the 95\% interval for the paired AUROC difference, from a bootstrap over papers with 2,000 resamples, is above $-3$, a margin we fixed before running the experiments.
Holistic judgment reads the same paper for every comment, so we count its repeated input at the cached price, which favors the baseline.

\section{Main Results}
\label{sec:results}

\subsection{Accuracy, cost, and latency}
\label{sec:main}
\label{sec:time}

\begin{wrapfigure}{r}{0.45\textwidth}
\centering
\vspace{-\baselineskip}\setlength{\abovecaptionskip}{3pt}
\includegraphics[width=\linewidth]{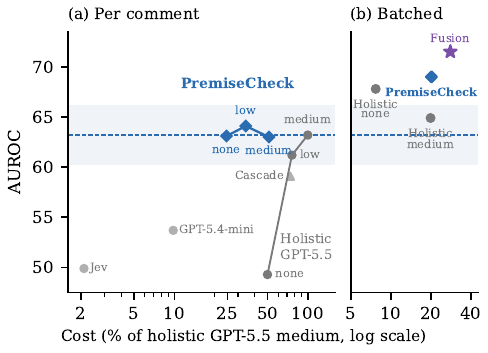}
\caption{Cost and AUROC on the NLP reviews, (a) per comment and (b) batched per review (\S\ref{sec:batch}), where \method{} uses 16 phrasings. Labels give the reasoning of the LLM, and the line and band mark the per-comment reference and $\pm 3$.}
\label{fig:frontier}
\end{wrapfigure}

Table~\ref{tab:main} and Figure~\ref{fig:frontier}a compare \method{} with holistic judgment and a cascade when each comment is checked on its own.
(1)~\textbf{\method{} matches holistic \gpt{} at about half the cost.}
With a medium-reasoning decomposer, it is within 0.3 of the reference or above it on all three parts of \bench{}, non-inferior on the NLP reviews, and not significantly different on the other two (Appendix~\ref{app:other-parts}).
(2)~\textbf{Without reasoning, the cost falls to a quarter on the NLP reviews}, where \method{} remains non-inferior (\S\ref{sec:reasoning}), while on the other two parts it loses 1.7 to 3.2 points, not significantly.
(3)~\textbf{Without reasoning, \method{} takes less than half the time.}
It needs 45\% of the time of holistic \gpt{} on the NLP reviews and 37\% and 22\% on the other parts (Appendix~\ref{app:latency}).
(4)~\textbf{Cheaper per-comment baselines lose accuracy.}
Holistic \jev{} and \mini{} fall 5.7 to 13.4 below the reference, holistic \gpt{} without reasoning falls 14.0 on the NLP reviews at twice the cost of \method{}, and the cascade costs about three quarters of the reference yet stays below \method{} with the medium-reasoning decomposer on every part.

\subsection{Robustness}
\label{sec:robust}

\begin{wrapfigure}{r}{0.45\textwidth}
\centering
\vspace{-\baselineskip}\setlength{\abovecaptionskip}{3pt}
\includegraphics[width=\linewidth]{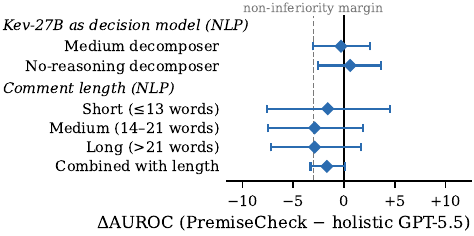}
\caption{AUROC of \method{} minus holistic \gpt{} on the NLP reviews, with 95\% intervals.}
\label{fig:robust}
\end{wrapfigure}

Figure~\ref{fig:robust} repeats the comparison with holistic \gpt{} on the NLP reviews in two further settings (Appendix~\ref{app:robust}).
(1)~\textbf{An open-weight decision model works as well.}
With \kev{} \citep{palmer2026kev27b} as router and verifier, \method{} reaches the same AUROC as with \jev{}, although with the medium-reasoning decomposer its interval falls just short of non-inferiority.
With the no-reasoning decomposer, it is non-inferior at an API cost of 17.8\% of the reference, not counting the GPU time of \kev{}.
(2)~\textbf{\method{} is not a proxy for comment length}, although longer comments are more often factual errors.
Within groups of similar length and after combining each score with length, \method{} stays within 3 points of holistic \gpt{} without a significant difference, though none of these comparisons reaches non-inferiority, and adding \method{} to length raises AUROC significantly, by 3.3.

\section{What Makes \method{} Work}
\label{sec:analysis}

\subsection{Decomposition makes decision models useful}
\label{sec:checker}
\label{sec:models}

\begin{figure}[t]
\centering
\begin{minipage}[t]{0.485\textwidth}
\centering
\includegraphics[width=\linewidth]{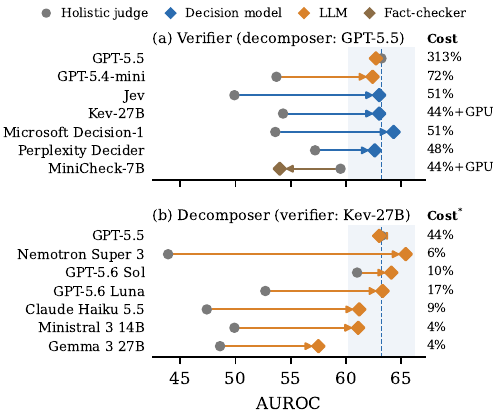}
\caption{AUROC on the NLP reviews of each model as a holistic judge and inside \method{}. Cost is relative to holistic \gpt{} in (a) and to the same LLM in (b), where $^{*}$ marks API cost. The band marks holistic \gpt{} $\pm 3$ (Appendices~\ref{app:checkers},~\ref{app:models}).}
\label{fig:models}
\end{minipage}\hfill
\begin{minipage}[t]{0.485\textwidth}
\centering
\includegraphics[width=\linewidth]{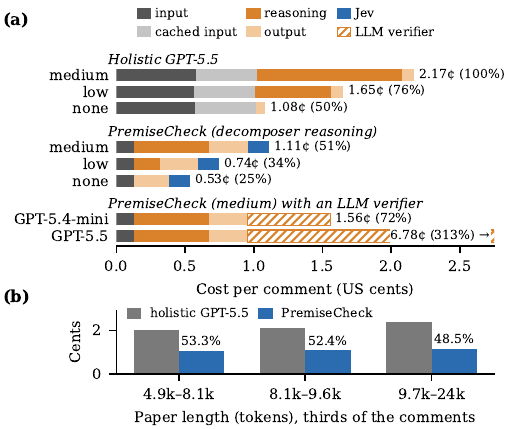}
\caption{Cost per comment on the NLP reviews (a) by \gpt{} token type and verifier and (b) by paper length at medium reasoning (Appendix~\ref{app:cost}).}
\label{fig:cost}
\end{minipage}
\end{figure}

\method{} differs from holistic \gpt{} in two ways: it decomposes the comment, and a decision model reads the paper.
Figure~\ref{fig:models} separates the two by scoring each model as a holistic judge and inside \method{} on the NLP reviews.
(1)~\textbf{Decomposition lets a cheap verifier match a frontier LLM.}
On the same premises, \gpt{} as the verifier is no more accurate than \jev{} but makes the pipeline cost 3.1 times as much as holistic judgment, and the \mini{} verifier makes it cost 1.4 times as much as with \jev{}.
(2)~\textbf{Every decision model gains from decomposition.}
\jev{}, \kev{}, Microsoft Decision-1, and Perplexity Decider score 50 to 57 as holistic judges but come within one point of holistic \gpt{} as verifiers, although the intervals of \kev{} and Perplexity Decider fall just short of non-inferiority, whereas the fact-checking model MiniCheck-7B \citep{tang2024minicheck} falls below its holistic score because it rejects many correct premises (Appendix~\ref{app:minicheck}).
(3)~\textbf{Much cheaper LLMs can decompose.}
With \kev{} as the verifier, GPT-5.6 Sol, GPT-5.6 Luna, and Nemotron Super 3 as decomposers are non-inferior to holistic \gpt{} at 6\% to 17\% of the cost of their holistic judgment, while weaker decomposers lose 2 to 6 points.

\subsection{Where the savings come from}
\label{sec:reasoning}
\label{sec:paperlen}

Figure~\ref{fig:cost} breaks down the cost on the NLP reviews.
(1)~\textbf{The decomposer needs no reasoning.}
Reasoning makes up about half the cost of holistic \gpt{}, and without it per-comment holistic judgment falls to chance level on the NLP reviews.
\method{} keeps its accuracy there without reasoning, because the LLM only rewrites the comment, while \jev{} decides whether the paper supports each premise.
At low or no reasoning, \method{} is thus more accurate than per-comment holistic judgment on the NLP reviews at less than half its cost.
(2)~\textbf{Longer papers yield larger savings.}
Only \jev{} reads the paper, so the relative cost of \method{} falls from 53\% for the shortest third of the papers to 49\% for the longest third.

\subsection{Design choices}
\label{sec:design}
\label{sec:swarm}

\begin{figure}[t]
\centering
\begin{minipage}[t]{0.485\textwidth}
\centering
\includegraphics[width=\linewidth]{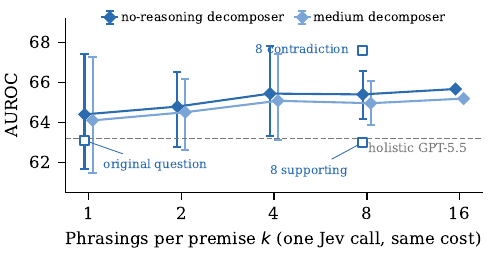}
\caption{AUROC on the NLP reviews by number of phrasings, as means over 50 random subsets of the 16 phrasings with 5--95\% ranges. Squares are fixed sets.}
\label{fig:swarm}
\end{minipage}\hfill
\begin{minipage}[t]{0.485\textwidth}
\centering
\includegraphics[width=\linewidth]{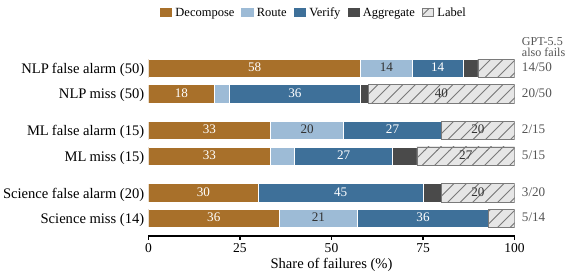}
\caption{Stage that causes each sampled failure of \method{}. False alarms are correct comments in the lowest 10\% of scores, and misses are errors above the median.}
\label{fig:failures}
\end{minipage}
\end{figure}

(1)~\textbf{Routing is needed, but the exact route is not.}
\kev{} agrees with \jev{} on only 65\% of the routes yet gives the same AUROC, whereas skipping routing lowers AUROC from 61.8 to 58.8 on a common set of machine learning premises, because knowledge premises then receive low probabilities (Appendix~\ref{app:design}).
(2)~\textbf{The weakest premise decides.}
Taking the mean over premises instead of the minimum lowers AUROC to 59.4, although the mean is slightly better on a label that also counts errors of judgment (Appendix~\ref{app:design}).
(3)~\textbf{Contradiction questions add accuracy at no cost.}
Because \jev{} is billed only for its input, we ask 16 phrasings of the verification question in one call and average them, eight asking whether the paper supports the premise and eight whether it contradicts the premise (Appendix~\ref{app:statements}).
Figure~\ref{fig:swarm} shows that AUROC levels off after four phrasings and rises from 63.1 to 65.7 with all 16, a significant gain over the original question.
The gain comes from the contradiction questions, since a supporting question also penalizes a premise that the paper does not mention.
Without development data on which to select a subset, we keep all 16.

\subsection{Error types and scope}
\label{sec:scope}

(1)~\textbf{\method{} catches both main types of factual error.}
On Misunderstanding and Neglect, which make up almost all errors in the NLP reviews, \method{} with 16 phrasings scores above holistic \gpt{}, though not significantly (Appendix~\ref{app:scope}).
For 79\% of the Neglect errors, its lowest premise claims an omission, compared with 36\% for correct comments.
(2)~\textbf{\method{} targets facts, not judgment.}
It treats a comment as wrong when one premise is false, which does not hold for errors of judgment.
On labels from \prb{} and \rs{} that also count unimportant concerns or unreasonable requests, \method{} falls behind holistic \gpt{}, significantly on \prb{} and by a wide margin on the few such errors of \rs{}, but on factual errors alone the gap disappears.
Appendix~\ref{app:examples} shows scored comments with their premises.

\subsection{Where \method{} fails}
\label{sec:failures}

Three annotators assign 164 sampled failures to the stage that causes them (Appendix~\ref{app:failures}).
Figure~\ref{fig:failures} shows that false alarms come mostly from decomposition on the NLP reviews (58\%), where a premise adds to, redirects, or blurs what the reviewer said, and mostly from verification on the natural-science reviews (45\%).
Holistic \gpt{} fails on only 13\% and 15\% of the machine learning and natural-science false alarms, so these are largely specific to \method{}.
Misses on the NLP reviews split between questionable labels (40\%) and verification (36\%), and holistic \gpt{} also fails on 12 of the 20 comments with a questionable label.

\section{Using \method{} in Practice}
\label{sec:application}

\begin{figure}[t]
\centering
\begin{minipage}[t]{0.54\textwidth}
\centering
\includegraphics[width=\linewidth]{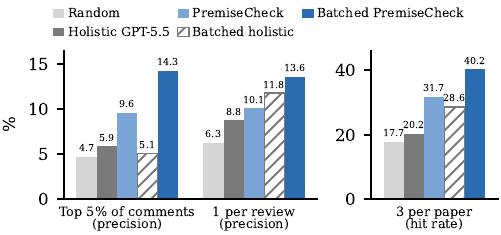}
\caption{Flagging factual errors on the NLP reviews. Hit rate is the share of papers with an error that one of the three flags hits. \method{} uses 16 phrasings, and ties are broken at random.}
\label{fig:flagging}
\end{minipage}\hfill
\begin{minipage}[t]{0.42\textwidth}
\centering
\includegraphics[width=\linewidth]{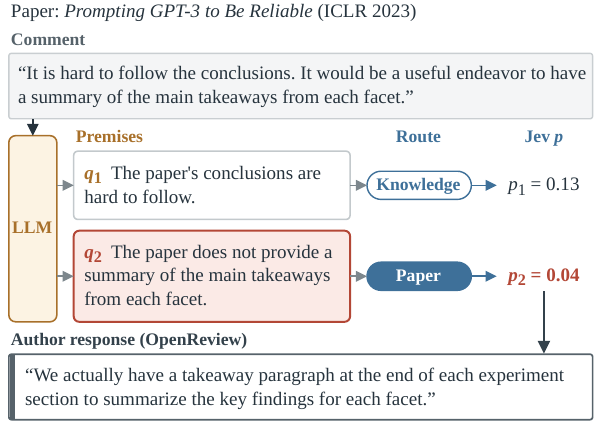}
\caption{A flagged comment from the machine learning reviews with the authors' OpenReview response, quoted verbatim.}
\label{fig:flag-example}
\end{minipage}
\end{figure}

\subsection{Batching and fusion}
\label{sec:batch}
\label{sec:fusion}

When a review is finished, as for an area chair or an author, all its comments can be checked in one call, as in the original \rc{} study \citep{du2024reviewcritique}.
We batch both approaches per review and fuse them by averaging the percentile ranks of their scores.
Figure~\ref{fig:frontier}b shows these systems, and Appendix~\ref{app:batch} gives the details.
(1)~\textbf{Batching helps both approaches.}
Batched holistic \gpt{} without reasoning reaches 67.8 at 7.7\% of the reference cost, and batched \method{} reaches 69.0 at 20\%.
The two do not differ significantly, so holistic \gpt{} is the cheaper choice here, while \method{} also returns the premises.
(2)~\textbf{Fusion gives the best result.}
The two scores have a rank correlation of only 0.56, and batched fusion reaches 71.5 at 28\% of the reference cost, significantly above batched holistic \gpt{}.

\subsection{Flagging comments for a person to check}
\label{sec:flagging}

A deployed system would flag a few comments for a person to check, each with the premise that \jev{} finds false.
Figure~\ref{fig:flagging} simulates three flagging policies on the NLP reviews, where 4.8\% of the comments contain a factual error.
(1)~\textbf{Flags of \method{} hit errors two to three times as often as random.}
This holds for \method{} and batched \method{} among the 5\% most suspicious comments, whereas holistic \gpt{}, per comment or batched, does little better than random despite a similar AUROC.
(2)~\textbf{An area chair sees more errors.}
Reading the three most suspicious comments of each paper finds an error for 40\% of the papers that contain one with batched \method{}, compared with 29\% with batched holistic \gpt{}.
(3)~\textbf{A flag points to the premise that the paper refutes.}
In Figure~\ref{fig:flag-example}, \jev{} gives the premise that the paper has no summary of its takeaways a probability of 0.04, and the authors reply that each experiment section ends with one.
Appendix~\ref{app:rebuttal} explains how we chose this comment and shows a second example and a false alarm.

\section{Conclusion}
\label{sec:conclusion}

Fact-checking review comments does not require a frontier LLM to read the paper for every comment.
\bench{} labels factual errors in review comments under one definition across three fields.
In \method{}, an LLM only rewrites each comment as explicit premises, and a cheap decision model checks them against the paper.
When each comment is checked on its own, this division of labor reaches accuracy comparable to \gpt{} on every part of \bench{} at a fraction of its cost.
\added{This result does not hinge on one model.
Decision models from several providers can verify the premises, and on the NLP reviews, LLMs much cheaper than \gpt{}, or \gpt{} without reasoning, can write them.}

\added{\method{} can support peer review at two points.
While a reviewer writes, it can check each comment as soon as the comment is written, which is the setting of our main experiments.
Once a review is finished, an area chair or an author can check all its comments together, where fusing \method{} with holistic \gpt{} gives the best accuracy.
In both cases, \method{} is meant to flag a few comments for a person and show the premise that the paper appears to refute, so that the person can check that one claim instead of rereading the paper.}

\added{Several directions remain open.
Most false alarms on the NLP reviews arise in decomposition, so a decomposer that stays closer to what the reviewer said is a natural next step.
Errors of judgment, such as an unimportant concern or an unreasonable request, lie outside \method{} and call for different methods.
Finally, \bench{} can measure future checkers on accuracy, cost, and latency together, and further open-weight decision models can be tested within the same framework.}

\section*{Limitations}
\method{} targets factual errors about the paper, that is, comments that can be checked against what the paper states.
A comment can also be wrong in its judgment, for example about the importance of a concern or the novelty of the work (\S\ref{sec:scope}).
Detecting such errors of judgment is a direction for future work.

Our main decision model, \jev{}, is served through an API.
With the open-weight \kev{} as the decision model, \method{} reaches the same AUROC on the NLP reviews of \bench{} (\S\ref{sec:robust}).
Testing further open decision models is left for future work.

\section*{Ethical Considerations}
We use \rc{}, \rs{}, and \prb{} for research evaluation only. The license of \rc{} restricts it to research and evaluation, and we do not train any model on it.
We release our code under the MIT license and the NLP and natural-science parts of \bench{} under the licenses of \rc{} and \prb{}, which allow evaluation but not training.
\rs{} states no license, so for the machine-learning part we release a script that rebuilds it from the original data instead of the data itself.
A verifier that wrongly flags a correct comment could unfairly discredit a reviewer, and one that misses an error could lend false authority to a wrong comment.
\method{} is meant to direct human attention to comments that may misstate the paper, not to replace the judgment of reviewers, area chairs, or authors.

\bibliography{refs}

\begin{thebibliography}{36}
\providecommand{\natexlab}[1]{#1}

\bibitem[{Almeida(2026)}]{typesafe2026blog}
Diogo Almeida. 2026.
\newblock \href
  {https://typesafe.ai/blog/introducing-system-one-models-and-jev} {Introducing
  system one models \& jev}.
\newblock TypeSafe AI Blog.

\bibitem[{Biswas et~al.(2026)Biswas, Schoepp, Vasan, Opipari, Zhang, Hu,
  Joseph, Lease, Li, Stone, Wagstaff, Taylor, and Jenkins}]{biswas2026aaai26}
Joydeep Biswas, Sheila Schoepp, Gautham Vasan, Anthony Opipari, Arthur Zhang,
  Zichao Hu, Sebastian Joseph, Matthew Lease, Junyi~Jessy Li, Peter Stone,
  Kiri~L. Wagstaff, Matthew~E. Taylor, and Odest~Chadwicke Jenkins. 2026.
\newblock \href {https://arxiv.org/abs/2604.13940} {{AI-Assisted Peer Review at
  Scale: The AAAI-26 AI Review Pilot}}.
\newblock \emph{arXiv preprint arXiv:2604.13940}.

\bibitem[{Chen et~al.(2024)Chen, Zaharia, and Zou}]{chen2023frugalgpt}
Lingjiao Chen, Matei Zaharia, and James Zou. 2024.
\newblock {FrugalGPT}: How to use large language models while reducing cost and
  improving performance.
\newblock \emph{Transactions on Machine Learning Research}.
\newblock ArXiv:2305.05176.

\bibitem[{D'Arcy et~al.(2024)D'Arcy, Hope, Birnbaum, and
  Downey}]{darcy2024marg}
Mike D'Arcy, Tom Hope, Larry Birnbaum, and Doug Downey. 2024.
\newblock \href {https://arxiv.org/abs/2401.04259} {{MARG}: Multi-agent review
  generation for scientific papers}.
\newblock \emph{arXiv preprint arXiv:2401.04259}.

\bibitem[{Dekoninck et~al.(2025)Dekoninck, Baader, and
  Vechev}]{dekoninck2025cascaderouting}
Jasper Dekoninck, Maximilian Baader, and Martin Vechev. 2025.
\newblock A unified approach to routing and cascading for {LLMs}.
\newblock In \emph{International Conference on Machine Learning (ICML)}.
\newblock ArXiv:2410.10347.

\bibitem[{Du et~al.(2024)Du, Wang, Zhao, Deng, Liu, Lou, Zou, Narayanan~Venkit,
  Zhang, Srinath, Zhang, Gupta, Li, Li, Wang, Liu, Liu, Gao, Xia, Xing,
  Jiayang, Wang, Su, Shah, Guo, Gu, Li, Wei, Wang, Cheng, Ranathunga, Fang, Fu,
  Liu, Huang, Blanco, Cao, Zhang, Yu, and Yin}]{du2024reviewcritique}
Jiangshu Du, Yibo Wang, Wenting Zhao, Zhongfen Deng, Shuaiqi Liu, Renze Lou,
  Henry~Peng Zou, Pranav Narayanan~Venkit, Nan Zhang, Mukund Srinath,
  Haoran~Ranran Zhang, Vipul Gupta, Yinghui Li, Tao Li, Fei Wang, Qin Liu,
  Tianlin Liu, Pengzhi Gao, Congying Xia, and 21 others. 2024.
\newblock \href {https://doi.org/10.18653/v1/2024.emnlp-main.292} {{LLM}s
  assist {NLP} researchers: Critique paper (meta-)reviewing}.
\newblock In \emph{Proceedings of EMNLP 2024}, pages 5081--5099.

\bibitem[{Erol et~al.(2026)Erol, El, Suzgun, Yuksekg{\"o}n{\"u}l, and
  Zou}]{erol2025costofpass}
Mehmet~Hamza Erol, Batu El, Mirac Suzgun, Mert Yuksekg{\"o}n{\"u}l, and James
  Zou. 2026.
\newblock Cost-of-pass: An economic framework for evaluating language models.
\newblock In \emph{International Conference on Learning Representations
  (ICLR)}.
\newblock ArXiv:2504.13359.

\bibitem[{Ghorbanpour et~al.(2026)Ghorbanpour, Sadeghian, Daghighfarsoodeh,
  Ebrahimi, Arabzadeh, Hosseini, and Bagheri}]{ghorbanpour2026peerispect}
Ali Ghorbanpour, Soroush Sadeghian, Alireza Daghighfarsoodeh, Sajad Ebrahimi,
  Negar Arabzadeh, Seyed~Mohammad Hosseini, and Ebrahim Bagheri. 2026.
\newblock \href {https://doi.org/10.1145/3805712.3808368} {Peerispect: Claim
  verification in scientific peer reviews}.
\newblock In \emph{Proceedings of the 49th International ACM SIGIR Conference
  on Research and Development in Information Retrieval}, pages 5161--5165.

\bibitem[{Jung et~al.(2025)Jung, Brahman, and Choi}]{jung2025trustorescalate}
Jaehun Jung, Faeze Brahman, and Yejin Choi. 2025.
\newblock Trust or escalate: {LLM} judges with provable guarantees for human
  agreement.
\newblock In \emph{International Conference on Learning Representations
  (ICLR)}.
\newblock ArXiv:2407.18370.

\bibitem[{Kapoor et~al.(2025)Kapoor, Stroebl, Siegel, Nadgir, and
  Narayanan}]{kapoor2024agentsmatter}
Sayash Kapoor, Benedikt Stroebl, Zachary~S. Siegel, Nitya Nadgir, and Arvind
  Narayanan. 2025.
\newblock {AI} agents that matter.
\newblock \emph{Transactions on Machine Learning Research}.
\newblock ArXiv:2407.01502.

\bibitem[{Khattab et~al.(2024)Khattab, Singhvi, Maheshwari, Zhang, Santhanam,
  Vardhamanan, Haq, Sharma, Joshi, Moazam, Miller, Zaharia, and
  Potts}]{khattab2023dspy}
Omar Khattab, Arnav Singhvi, Paridhi Maheshwari, Zhiyuan Zhang, Keshav
  Santhanam, Sri Vardhamanan, Saiful Haq, Ashutosh Sharma, Thomas~T. Joshi,
  Hanna Moazam, Heather Miller, Matei Zaharia, and Christopher Potts. 2024.
\newblock {DSPy}: Compiling declarative language model calls into
  state-of-the-art pipelines.
\newblock In \emph{International Conference on Learning Representations
  (ICLR)}.
\newblock ArXiv:2310.03714.

\bibitem[{Kim et~al.(2025)Kim, Lee, and Lee}]{kim2025reviewcrisis}
Jaeho Kim, Yunseok Lee, and Seulki Lee. 2025.
\newblock \href {https://arxiv.org/abs/2505.04966} {Position: The {AI}
  conference peer review crisis demands author feedback and reviewer rewards}.
\newblock In \emph{International Conference on Machine Learning (ICML),
  Position Paper Track}.
\newblock ArXiv:2505.04966.

\bibitem[{Kim et~al.(2026)Kim, Yoon, Gashteovski, Suk, Baek, Aggarwal, Wu,
  Zaverkin, Petkoski, Schrider, Dukovski, Santini, Mitreska, Jeong, Kwon, Sim,
  Manasova, Porto, Mojsoska, Takamoto, Shuntov, Liu, Lee, Din{\c{c}}, Jo, Han,
  Lee, Li, Tsai, Simsek, Shafi, Chung, Park, Shulevski, Christiansen, Son,
  Knight, Montoya, Ahn, Langkammer, Moon, Yoon, Stikov, Jang, Choi, Kim, Jung,
  Kim, Kim, Anjum, Kim, Bridges, Lawrence, Yue, Oh, Asai, Welleck, and
  Neubig}]{kim2026peerreviewbench}
Seungone Kim, Dongkeun Yoon, Kiril Gashteovski, Juyoung Suk, Jinheon Baek,
  Pranjal Aggarwal, Ian Wu, Viktor Zaverkin, Spase Petkoski, Daniel~R.
  Schrider, Ilija Dukovski, Francesco Santini, Biljana Mitreska, Yong Jeong,
  Kyeongha Kwon, Young~Min Sim, Dragana Manasova, Arthur Porto, Biljana
  Mojsoska, and 39 others. 2026.
\newblock \href {https://arxiv.org/abs/2605.20668} {On the limits and
  opportunities of {AI} reviewers: Reviewing the reviews of {Nature}-family
  papers with 45 expert scientists}.
\newblock \emph{arXiv preprint arXiv:2605.20668}.

\bibitem[{Kuznetsov et~al.(2024)Kuznetsov, Afzal, Dercksen, Dycke, Goldberg,
  Hope, Hovy, Kummerfeld, Lauscher, Leyton-Brown, Lu, {Mausam}, Mieskes,
  N{\'e}v{\'e}ol, Pruthi, Qu, Schwartz, Smith, Solorio, Wang, Zhu, Rogers,
  Shah, and Gurevych}]{kuznetsov2024nlp}
Ilia Kuznetsov, Osama~Mohammed Afzal, Koen Dercksen, Nils Dycke, Alexander
  Goldberg, Tom Hope, Dirk Hovy, Jonathan~K. Kummerfeld, Anne Lauscher, Kevin
  Leyton-Brown, Sheng Lu, {Mausam}, Margot Mieskes, Aur{\'e}lie N{\'e}v{\'e}ol,
  Danish Pruthi, Lizhen Qu, Roy Schwartz, Noah~A. Smith, Thamar Solorio, and 5
  others. 2024.
\newblock \href {https://arxiv.org/abs/2405.06563} {What can natural language
  processing do for peer review?}
\newblock \emph{arXiv preprint arXiv:2405.06563}.

\bibitem[{Laban et~al.(2022)Laban, Schnabel, Bennett, and
  Hearst}]{laban2022summac}
Philippe Laban, Tobias Schnabel, Paul~N. Bennett, and Marti~A. Hearst. 2022.
\newblock \href {https://doi.org/10.1162/tacl_a_00453} {{SummaC}: Re-visiting
  {NLI}-based models for inconsistency detection in summarization}.
\newblock \emph{Transactions of the Association for Computational Linguistics},
  10:163--177.

\bibitem[{Liang et~al.(2024{\natexlab{a}})Liang, Izzo, Zhang, Lepp, Cao, Zhao,
  Chen, Ye, Liu, Huang, McFarland, and Zou}]{liang2024monitoring}
Weixin Liang, Zachary Izzo, Yaohui Zhang, Haley Lepp, Hancheng Cao, Xuandong
  Zhao, Lingjiao Chen, Haotian Ye, Sheng Liu, Zhi Huang, Daniel~A. McFarland,
  and James~Y. Zou. 2024{\natexlab{a}}.
\newblock \href {https://arxiv.org/abs/2403.07183} {Monitoring {AI}-modified
  content at scale: A case study on the impact of {ChatGPT} on {AI} conference
  peer reviews}.
\newblock In \emph{International Conference on Machine Learning (ICML)}.
\newblock ArXiv:2403.07183.

\bibitem[{Liang et~al.(2024{\natexlab{b}})Liang, Zhang, Cao, Wang, Ding, Yang,
  Vodrahalli, He, Smith, Yin, McFarland, and Zou}]{liang2024feedback}
Weixin Liang, Yuhui Zhang, Hancheng Cao, Binglu Wang, Daisy Ding, Xinyu Yang,
  Kailas Vodrahalli, Siyu He, Daniel Smith, Yian Yin, Daniel McFarland, and
  James Zou. 2024{\natexlab{b}}.
\newblock \href {https://arxiv.org/abs/2310.01783} {Can large language models
  provide useful feedback on research papers? {A} large-scale empirical
  analysis}.
\newblock \emph{NEJM AI}, 1(8).
\newblock ArXiv:2310.01783.

\bibitem[{Min et~al.(2023)Min, Krishna, Lyu, Lewis, Yih, Koh, Iyyer,
  Zettlemoyer, and Hajishirzi}]{min2023factscore}
Sewon Min, Kalpesh Krishna, Xinxi Lyu, Mike Lewis, Wen-tau Yih, Pang~Wei Koh,
  Mohit Iyyer, Luke Zettlemoyer, and Hannaneh Hajishirzi. 2023.
\newblock {FActScore}: Fine-grained atomic evaluation of factual precision in
  long form text generation.
\newblock In \emph{Proceedings of the 2023 Conference on Empirical Methods in
  Natural Language Processing}.

\bibitem[{Ong et~al.(2025)Ong, Almahairi, Wu, Chiang, Wu, Gonzalez, Kadous, and
  Stoica}]{ong2025routellm}
Isaac Ong, Amjad Almahairi, Vincent Wu, Wei-Lin Chiang, Tianhao Wu, Joseph~E.
  Gonzalez, M~Waleed Kadous, and Ion Stoica. 2025.
\newblock {RouteLLM}: Learning to route {LLMs} from preference data.
\newblock In \emph{International Conference on Learning Representations
  (ICLR)}.
\newblock ArXiv:2406.18665.

\bibitem[{Ou et~al.(2025)Ou, Walden, Sanders, Jiang, Sun, Cheng, Jurayj,
  Wanner, Liang, Morgan, Han, Wang, May, Recknor, Khashabi, and
  Van~Durme}]{ou2025claimcheck}
Jiefu Ou, William Walden, Kate Sanders, Zhengping Jiang, Kaiser Sun, Jeffrey
  Cheng, William Jurayj, Miriam Wanner, Shaobo Liang, Candice Morgan, Seunghoon
  Han, Weiqi Wang, Chandler May, Hannah Recknor, Daniel Khashabi, and Benjamin
  Van~Durme. 2025.
\newblock {CLAIMCHECK}: How grounded are {LLM} critiques of scientific papers?
\newblock In \emph{Findings of the Association for Computational Linguistics:
  EMNLP 2025}, pages 21712--21735.

\bibitem[{Palmer(2026)}]{palmer2026kev27b}
Jared Palmer. 2026.
\newblock {Kev-27B}.
\newblock \url{https://huggingface.co/jaredpalmer/kev-27b}.
\newblock Hugging Face model card, Kev 1.0, fine-tuned from Qwen3.8-27B (Hub
  commit af0e6d5).

\bibitem[{Ryu et~al.(2025)Ryu, Jang, Lee, Jeong, Kim, Cho, Chu, Hwang, Jang,
  Kim, Kim, Kim, Kim, Kim, Lee, Park, Yun, Betz, and Yang}]{ryu2025reviewscore}
Hyun Ryu, Doohyuk Jang, Hyemin~S. Lee, Joonhyun Jeong, Gyeongman Kim, Donghyeon
  Cho, Gyouk Chu, Minyeong Hwang, Hyeongwon Jang, Changhun Kim, Haechan Kim,
  Jina Kim, Joowon Kim, Yoonjeon Kim, Kwanhyung Lee, Chanjae Park, Heecheol
  Yun, Gregor Betz, and Eunho Yang. 2025.
\newblock {ReviewScore}: Misinformed peer review detection with large language
  models.
\newblock \emph{arXiv preprint arXiv:2509.21679}.

\bibitem[{Shah(2022)}]{shah2022challenges}
Nihar~B. Shah. 2022.
\newblock \href {https://doi.org/10.1145/3528086} {Challenges, experiments, and
  computational solutions in peer review}.
\newblock \emph{Communications of the ACM}, 65(6):76--87.

\bibitem[{Snell et~al.(2025)Snell, Lee, Xu, and Kumar}]{snell2025scaling}
Charlie Snell, Jaehoon Lee, Kelvin Xu, and Aviral Kumar. 2025.
\newblock \href {https://arxiv.org/abs/2408.03314} {Scaling {LLM} test-time
  compute optimally can be more effective than scaling parameters for
  reasoning}.
\newblock In \emph{International Conference on Learning Representations
  (ICLR)}.
\newblock ArXiv:2408.03314.

\bibitem[{Son et~al.(2025)Son, Hong, Fan, Nam, Ko, Lim, Song, Choi, Paulo, Yu,
  and Biderman}]{son2025spot}
Guijin Son, Jiwoo Hong, Honglu Fan, Heejeong Nam, Hyunwoo Ko, Seungwon Lim,
  Jinyeop Song, Jinha Choi, Gon{\c{c}}alo Paulo, Youngjae Yu, and Stella
  Biderman. 2025.
\newblock \href {https://arxiv.org/abs/2505.11855} {When {AI} co-scientists
  fail: {SPOT}---a benchmark for automated verification of scientific
  research}.
\newblock \emph{arXiv preprint arXiv:2505.11855}.

\bibitem[{Tang et~al.(2024)Tang, Laban, and Durrett}]{tang2024minicheck}
Liyan Tang, Philippe Laban, and Greg Durrett. 2024.
\newblock {MiniCheck}: Efficient fact-checking of {LLM}s on grounding
  documents.
\newblock In \emph{Proceedings of EMNLP}.

\bibitem[{Thakkar et~al.(2025)Thakkar, Yuksekgonul, Silberg, Garg, Peng, Sha,
  Yu, Vondrick, and Zou}]{thakkar2025reviewfeedback}
Nitya Thakkar, Mert Yuksekgonul, Jake Silberg, Animesh Garg, Nanyun Peng, Fei
  Sha, Rose Yu, Carl Vondrick, and James Zou. 2025.
\newblock \href {https://arxiv.org/abs/2504.09737} {Can {LLM} feedback enhance
  review quality? {A} randomized study of 20{K} reviews at {ICLR} 2025}.
\newblock \emph{arXiv preprint arXiv:2504.09737}.

\bibitem[{{TypeSafe AI}(2026)}]{typesafe2026docs}
{TypeSafe AI}. 2026.
\newblock {TypeSafe AI} documentation: Models ({Jev 1.13}), primitives, and
  model jaggedness.
\newblock \url{https://docs.typesafe.ai/}.
\newblock Model documentation for Jev 1.13.

\bibitem[{Wang et~al.(2023)Wang, Wei, Schuurmans, Le, Chi, Narang, Chowdhery,
  and Zhou}]{wang2023selfconsistency}
Xuezhi Wang, Jason Wei, Dale Schuurmans, Quoc Le, Ed~Chi, Sharan Narang,
  Aakanksha Chowdhery, and Denny Zhou. 2023.
\newblock \href {https://arxiv.org/abs/2203.11171} {Self-consistency improves
  chain of thought reasoning in language models}.
\newblock In \emph{International Conference on Learning Representations
  (ICLR)}.
\newblock ArXiv:2203.11171.

\bibitem[{Wei et~al.(2024)Wei, Yang, Song, Lu, Hu, Huang, Tran, Peng, Liu,
  Huang, Du, and Le}]{wei2024safe}
Jerry Wei, Chengrun Yang, Xinying Song, Yifeng Lu, Nathan Hu, Jie Huang, Dustin
  Tran, Daiyi Peng, Ruibo Liu, Da~Huang, Cosmo Du, and Quoc~V. Le. 2024.
\newblock Long-form factuality in large language models.
\newblock In \emph{Advances in Neural Information Processing Systems}.

\bibitem[{Zaharia et~al.(2024)Zaharia, Khattab, Chen, Davis, Miller, Potts,
  Zou, Carbin, Frankle, Rao, and Ghodsi}]{zaharia2024compound}
Matei Zaharia, Omar Khattab, Lingjiao Chen, Jared~Quincy Davis, Heather Miller,
  Chris Potts, James Zou, Michael Carbin, Jonathan Frankle, Naveen Rao, and Ali
  Ghodsi. 2024.
\newblock \href
  {https://bair.berkeley.edu/blog/2024/02/18/compound-ai-systems/} {The shift
  from models to compound {AI} systems}.
\newblock Berkeley Artificial Intelligence Research (BAIR) Blog.

\bibitem[{Zha et~al.(2023)Zha, Yang, Li, and Hu}]{zha2023alignscore}
Yuheng Zha, Yichi Yang, Ruichen Li, and Zhiting Hu. 2023.
\newblock {AlignScore}: Evaluating factual consistency with a unified alignment
  function.
\newblock In \emph{Proceedings of ACL}.

\bibitem[{Zhang and Abernethy(2025)}]{zhang2025critical}
Tianmai~M. Zhang and Neil~F. Abernethy. 2025.
\newblock \href {https://arxiv.org/abs/2505.23824} {Reviewing scientific papers
  for critical problems with reasoning {LLMs}: Baseline approaches and
  automatic evaluation}.
\newblock In \emph{NeurIPS 2025 Workshop on AI for Science}.
\newblock ArXiv:2505.23824.

\bibitem[{Zheng et~al.(2023)Zheng, Chiang, Sheng, Zhuang, Wu, Zhuang, Lin, Li,
  Li, Xing, Zhang, Gonzalez, and Stoica}]{zheng2023judging}
Lianmin Zheng, Wei-Lin Chiang, Ying Sheng, Siyuan Zhuang, Zhanghao Wu, Yonghao
  Zhuang, Zi~Lin, Zhuohan Li, Dacheng Li, Eric~P. Xing, Hao Zhang, Joseph~E.
  Gonzalez, and Ion Stoica. 2023.
\newblock \href {https://arxiv.org/abs/2306.05685} {Judging {LLM}-as-a-judge
  with {MT-Bench} and {Chatbot Arena}}.
\newblock In \emph{Advances in Neural Information Processing Systems (NeurIPS),
  Datasets and Benchmarks Track}.
\newblock ArXiv:2306.05685.

\bibitem[{Zhu et~al.(2025)Zhu, Weng, Yang, and Zhang}]{zhu2025deepreview}
Minjun Zhu, Yixuan Weng, Linyi Yang, and Yue Zhang. 2025.
\newblock \href {https://aclanthology.org/2025.acl-long.1420} {{DeepReview}:
  Improving {LLM}-based paper review with human-like deep thinking process}.
\newblock In \emph{Proceedings of the 63rd Annual Meeting of the Association
  for Computational Linguistics (ACL)}.
\newblock ArXiv:2503.08569.

\bibitem[{Zhuang et~al.(2025)Zhuang, Chen, Xu, Jiang, and
  Lin}]{zhuang2025survey}
Zhenzhen Zhuang, Jiandong Chen, Hongfeng Xu, Yuwen Jiang, and Jialiang Lin.
  2025.
\newblock \href {https://doi.org/10.1016/j.inffus.2025.103332} {Large language
  models for automated scholarly paper review: A survey}.
\newblock \emph{Information Fusion}, 124:103332.

\end{thebibliography}

\clearpage
\appendix
\etocdepthtag.toc{mtappendix}
\renewcommand{\topfraction}{0.95}
\renewcommand{\dbltopfraction}{0.96}
\renewcommand{\textfraction}{0.03}
\renewcommand{\floatpagefraction}{0.85}
\renewcommand{\dblfloatpagefraction}{0.85}
\setcounter{topnumber}{3}
\setcounter{dbltopnumber}{3}
\setcounter{totalnumber}{5}
\makeatletter
\setlength{\@fptop}{0pt}
\setlength{\@dblfptop}{0pt}
\setlength{\dblfloatsep}{8pt plus 2pt minus 2pt}
\setlength{\dbltextfloatsep}{14pt plus 2pt minus 4pt}
\setlength{\floatsep}{8pt plus 2pt minus 2pt}  %
\setlength{\textfloatsep}{14pt plus 2pt minus 4pt}
\makeatother

\etocsettagdepth{mtmain}{none}
\etocsettagdepth{mtappendix}{subsection}
\etocsettocstyle{\section*{Appendix Contents}}{}
\tableofcontents

\section{Benchmark Details}
\label{app:bench}

\subsection{Source Annotations}
\label{app:bench-annot}

All labels of \bench{} start from expert judgments in the three source datasets (\S\ref{sec:bench-data}).

\paragraph{\rc{} (NLP reviews).}
In \rc{} \citep{du2024reviewcritique}, 40 members of the NLP community, each with multiple first-authored papers at top NLP venues, label every sentence of a review as reliable or deficient and explain each deficiency.
They read the paper, all reviews, the meta-review, and the author rebuttal before labeling.
Two of them label each paper independently, and an expert with area-chair experience resolves disagreements.
Each labeled unit of a review is one comment in \bench{}.
Deficient comments carry fine-grained error types, four of which are the factual types of \bench{}.
Reliable comments are correct, and deficient comments with one of these four types are errors.
We leave out other deficiencies, such as an unconvincing argument, comments that only summarize the paper or list its strengths, and comments of fewer than four words, mostly headers such as ``Weaknesses:''.

\paragraph{\rs{} (machine learning reviews).}
In \rs{} \citep{ryu2025reviewscore}, 15 graduate students in AI rate each premise of an automatically reconstructed argument on a five-point factuality scale, and the median of three annotators is the final score.
Following \rs{}, an argument is an error when its lowest premise rating is below 3 and correct when it is above 3.
We leave out arguments at exactly 3, as well as claims and questions, which have no premise ratings.

\paragraph{\prb{} (natural-science reviews).}
In \prb{} \citep{kim2026peerreviewbench}, 45 scientists in the physical, biological, and health sciences judge whether each comment on a Nature-family paper is correct and may add a written rationale.
Comments judged correct are correct.
Comments judged incorrect are errors only when the error is factual, as decided in Appendix~\ref{app:bench-verify}.

\subsection{Label Verification}
\label{app:bench-verify}

\paragraph{Human verification.}
Two of the authors repeat the \gpt{} decision of \S\ref{sec:bench-data} on the 178 comments that \prb{} experts judged incorrect with a rationale.
Each annotator sees the comment, the expert rationale, and the paper text, but not the \gpt{} label, and the two annotators see the comments in different random orders.
The guidelines in Figure~\ref{fig:annot-guide} define the three labels with the error types of \rc{}.
Table~\ref{tab:annot-agree} reports the agreement.
Of the 38 comments with different labels, 17 are labeled factual by one annotator and judgment by the other, and 21 involve the unclear label.

\begin{figure}[!ht]
\begin{tcolorbox}[colback=white,colframe=rowgray,boxrule=0.4pt,arc=2pt,left=4pt,right=4pt,top=3pt,bottom=3pt,fontupper=\scriptsize]
Each row is a review comment that an expert annotator of \prb{} judged \textbf{Not Correct}, with the expert's rationale. Decide \textbf{why} the expert judged it incorrect, using the comment and the rationale. Open the paper only when the rationale is not enough to decide. Work alone, and do not discuss rows with the other annotator before both sheets are done. Choose exactly one label.
\begin{itemize}[leftmargin=10pt]
\item \textbf{factual}: the comment says something about \emph{this paper} that the paper does not support. Give the subtype.
  \begin{itemize}[leftmargin=10pt]
  \item \emph{Misunderstanding}: misreads what the paper does or claims.
  \item \emph{Neglect}: says the paper lacks something that it clearly contains (an experiment, a detail, a discussion).
  \item \emph{Unstated statement}: asserts something about the paper that has no basis in it.
  \item \emph{Inaccurate summary}: misstates the content or contribution of the paper.
  \end{itemize}
\item \textbf{judgment}: the comment is wrong in its judgment, not its facts, for example about importance, scope, or novelty, an unreasonable request, being too harsh or overstated, or a disagreement about field knowledge. An error about a \emph{cited} paper rather than this paper also counts as judgment.
\item \textbf{unclear}: the rationale does not let you decide.
\end{itemize}
When the rationale mainly says that the comment is too harsh or overstated but also mentions the paper's content, choose judgment.
\end{tcolorbox}
\caption{Guidelines for the human verification of the natural-science labels.}
\label{fig:annot-guide}
\end{figure}

\begin{table}[!ht]
\begin{minipage}[t]{0.47\textwidth}
\vspace{0pt}
\centering
\footnotesize
\setlength{\tabcolsep}{5pt}
\begin{tabular}{@{}lcccc@{}}
\toprule
 & \multicolumn{3}{c}{Annotator 2} & \\
\cmidrule(lr){2-4}
Annotator 1 & Factual & Judgment & Unclear & Total \\
\midrule
Factual & \cellcolor{okcell}47 & 12 & 6 & 65 \\
Judgment & 5 & \cellcolor{okcell}70 & 5 & 80 \\
Unclear & 1 & 9 & \cellcolor{okcell}23 & 33 \\
\midrule
Total & 53 & 91 & 34 & 178 \\
\midrule
Cohen's $\kappa$ & & & Binary & Three-way \\
\midrule
\multicolumn{3}{@{}l}{Annotator 1 vs.\ Annotator 2} & 0.70 & 0.66 \\
\multicolumn{3}{@{}l}{Annotator 1 vs.\ \gpt{}} & 0.76 & 0.72 \\
\multicolumn{3}{@{}l}{Annotator 2 vs.\ \gpt{}} & 0.77 & 0.77 \\
\bottomrule
\end{tabular}
\caption{Human verification of the natural-science labels on 178 comments. Top: labels of the two annotators, with agreement shaded. Bottom: Cohen's $\kappa$ for factual vs.\ other (binary) and for the three labels. \gpt{} labels 65 comments as factual, 92 as judgment, and 21 as unclear.}
\label{tab:annot-agree}
\end{minipage}\hfill
\begin{minipage}[t]{0.49\textwidth}
\vspace{0pt}
\centering
\footnotesize
\setlength{\tabcolsep}{2pt}
\begin{adjustbox}{max width=\linewidth}
\begin{tabular}{@{}lcccc@{}}
\toprule
Labels & Errors & \begin{tabular}[c]{@{}c@{}}Holistic\\\gpt{}\end{tabular} & \methodhd{} & $\Delta$ [95\% CI] \\
\midrule
\rowcolor{ourrow} \gpt{} (main) & 65 & 67.8 & 67.5 & $-0.3$ [$-8.9$, $+7.6$] \\
Annotator 1 & 65 & 64.6 & 62.4 & $-2.2$ [$-12.3$, $+7.2$] \\
Annotator 2 & 53 & 69.0 & 69.1 & $+0.2$ [$-9.4$, $+9.0$] \\
Both annotators & 47 & 67.1 & 69.8 & $+2.7$ [$-6.9$, $+11.6$] \\
Either annotator & 71 & 66.2 & 62.5 & $-3.7$ [$-13.8$, $+5.4$] \\
\bottomrule
\end{tabular}
\end{adjustbox}
\caption{AUROC on the natural-science reviews with factual labels from \gpt{} or from the annotators. ``Both'' and ``Either'' count a comment as a factual error when both or either annotator labels it factual. \method{} uses the medium-reasoning decomposer, and $\Delta$ is the paired difference to holistic \gpt{}.}
\label{tab:human-labels}
\end{minipage}
\end{table}

\paragraph{Results with human labels.}
Table~\ref{tab:human-labels} scores the natural-science reviews with the labels of the annotators in place of the \gpt{} labels on these 178 comments.
As with the \gpt{} labels, comments labeled judgment or unclear are left out.
The difference between \method{} and holistic \gpt{} ranges from $-3.7$ to $+2.7$, and no interval excludes zero.
\method{} is 2.7 above holistic \gpt{} on the 47 errors that both annotators label factual.

\paragraph{Automatic labels.}
\gpt{} at medium reasoning labels these comments with the following system prompt.
The prompt uses the definitions of Figure~\ref{fig:annot-guide} but shows only the comment and the expert rationale, not the paper.
The user message gives the comment, the rationale of the primary \prb{} expert, and the rationale of a second expert when that expert also judged the comment incorrect.

\begin{nolinenumbers}
\begin{promptbox}{Factual filter (natural-science labels), system prompt}
You classify why an expert judged a peer-review comment on a scientific paper to be incorrect.

You receive one review comment and the rationale of the expert annotator who judged it Not Correct (and, if a second annotator also judged it Not Correct, that annotator's rationale). You do not see the paper. Decide from the comment and the rationale alone.

Choose exactly one label:
- "factual": the comment says something about THIS paper that the paper does not support. Give one subtype:
  - "Misunderstanding": the comment misreads what the paper does or claims.
  - "Neglect": the comment says the paper lacks something that it clearly contains (an experiment, a detail, a discussion).
  - "Unstated statement": the comment asserts something about the paper that has no basis in it.
  - "Inaccurate Summary": the comment misstates the content or contribution of the paper.
- "judgment": the comment is wrong in its judgment, not its facts, for example about importance, scope or novelty, an unreasonable request, being too harsh or overstated, or a disagreement about field knowledge.
- "unclear": the rationale does not let you decide.

Rules:
- An error about a cited paper rather than this paper is "judgment".
- When the rationale mainly says the comment is too harsh or overstated but also mentions the paper's content, choose "judgment".
- Give the subtype only for "factual"; otherwise use null.
- Give a one-sentence reason.

Return ONLY JSON: {"label": "...", "subtype": "..." or null, "reason": "..."}
\end{promptbox}
\end{nolinenumbers}

\FloatBarrier
\section{Implementation Details}
\label{app:impl}

\subsection{Prompts}
\label{app:prompts}

\paragraph{Decomposition prompt.}
On the NLP reviews, the decomposer uses the following system prompt, and the user message is the review comment.

\begin{table}[t]
\centering
\footnotesize
\setlength{\tabcolsep}{4pt}
\begin{tabular}{@{}l>{\raggedright\arraybackslash}p{0.68\textwidth}@{}}
\toprule
Step & Question \\
\midrule
Route & Whether this statement is true can be decided from the content of the paper it refers to alone, without outside knowledge. \\
Verify (paper) & The proposition is true according to the paper. \\
Verify (knowledge) & The proposition is true. \\
Verify (natural science)$^\ast$ & The paper contains information that contradicts the proposition. \\
\midrule
Holistic & The review item's main criticism is factually correct about the paper. \\
\bottomrule
\end{tabular}
\caption{Questions given to the decision model. $^\ast$: inverted (one minus the probability), for paper-grounded premises on the natural-science reviews.}
\label{tab:questions}
\end{table}

\begin{nolinenumbers}
\begin{promptbox}{Decomposition, NLP reviews, system prompt}
You reconstruct ONE peer-review item as an argument, so that each step can be fact-checked separately.

List the premises the item relies on, explicit and implicit, and its conclusion.
- Each premise is ONE self-contained declarative sentence that is either true or false.
- Write it so it can be understood without the review item: say "the paper", and name the method, dataset, table, figure or equation it is about.
- Include premises about what the paper states, reports, shows or omits, and the general-knowledge premises that the reasoning needs.
- Keep each premise as close to the reviewer's own claim as possible. Do not add facts the reviewer did not assert or imply. Do not judge whether a premise is true.
- If the item is a question, write the presuppositions of the question as premises (for example what the reviewer assumes the paper does or does not state).
- Use 1 to 6 premises.

Return ONLY JSON: {"premises": ["...", "..."], "conclusion": "..."}
\end{promptbox}
\end{nolinenumbers}

\paragraph{Variant for the machine learning reviews.}
On the machine learning reviews, the decomposer keeps fewer premises.

\begin{nolinenumbers}
\begin{promptbox}{Decomposition, machine learning reviews, system prompt}
You reconstruct ONE peer-review item as a short argument, so that each step can be fact-checked separately.

List ONLY the premises the reviewer actually asserts or directly presupposes, in the reviewer's own terms:
- the reviewer's claims about what the paper states, reports, shows or omits;
- at most ONE general-knowledge premise, and only if the reviewer's conclusion depends on it.
Each premise is ONE self-contained declarative sentence: say "the paper" and name the method, dataset, table, figure or equation it is about.
Do NOT describe the paper beyond what the reviewer says. Do NOT add background facts, assumptions or inferences of your own. Do not judge whether a premise is true.
If the item is a question, write only what the question presupposes about the paper.
Use 1 to 4 premises.

Return ONLY JSON: {"premises": ["...", "..."], "conclusion": "..."}
\end{promptbox}
\end{nolinenumbers}

\begin{table}[t]
\centering
\footnotesize
\setlength{\tabcolsep}{3pt}
\begin{tabular}{@{}r>{\raggedright\arraybackslash}p{0.51\textwidth}>{\raggedright\arraybackslash}p{0.42\textwidth}@{}}
\toprule
$i$ & Paper-grounded premise & Knowledge premise \\
\midrule
0 & The proposition is true according to the paper. & The proposition is true. \\
1 & The paper states or clearly implies the proposition. & The proposition is correct. \\
2 & A careful reader of the paper would accept the proposition as an accurate description of the paper. & The proposition is a true statement. \\
3 & The proposition accurately describes the paper. & An expert in the field would agree with the proposition. \\
4 & The paper supports the proposition. & The proposition is accurate. \\
5 & The proposition is consistent with what the paper reports. & The proposition is factually correct. \\
6 & Someone who has read the paper closely would agree with the proposition. & The proposition reflects established knowledge in the field. \\
7 & The proposition is a faithful statement about the content of the paper. & The proposition is a well-supported claim. \\
\midrule
8$^\ast$ & The paper contains information that contradicts the proposition. & The proposition is false. \\
9$^\ast$ & The proposition misrepresents the paper. & The proposition is incorrect. \\
10$^\ast$ & The proposition is false according to the paper. & An expert in the field would reject the proposition. \\
11$^\ast$ & The proposition misreads or overlooks something that the paper states. & The proposition contains a factual error. \\
12$^\ast$ & The paper reports something that shows the proposition is wrong. & The proposition is inaccurate. \\
13$^\ast$ & A careful reader of the paper would reject the proposition. & The proposition is misleading. \\
14$^\ast$ & The proposition is inaccurate as a description of the paper. & The proposition contradicts established knowledge in the field. \\
15$^\ast$ & The proposition conflicts with the content of the paper. & The proposition is wrong. \\
\bottomrule
\end{tabular}
\caption{The 16 phrasings of the verification question. $^\ast$: inverted (one minus the probability).}
\label{tab:swarm-q}
\end{table}

\begin{table*}[t]
\centering
\small
\setlength{\tabcolsep}{5pt}
\begin{tabular}{@{}llccccc@{}}
\toprule
 & & & \multicolumn{2}{c}{Cost / comment} & & \\
\cmidrule(lr){4-5}
System & Reasoning & AUROC & \$ & \% & Latency (s) & $\Delta$ AUROC [95\% CI] \\
\midrule
\multicolumn{7}{@{}l}{\emph{Holistic judgment}} \\
\gpt{} (ref.) & medium & 63.2 & \$0.0217 & 100.0 & 6.27 & -- \\
\dimrow{\gpt{}} & \dimrow{low} & \dimrow{61.2} & \dimrow{\$0.0165} & \dimrow{76.1} & \dimrow{4.23} & \dimrow{$-2.1$ [$-3.1$, $-0.9$]} \\
\dimrow{\gpt{}} & \dimrow{none} & \dimrow{49.3} & \dimrow{\$0.0108} & \dimrow{49.9} & \dimrow{1.78} & \dimrow{$-14.0$ [$-16.7$, $-11.1$]} \\
\dimrow{\mini{}} & \dimrow{low} & \dimrow{53.7} & \dimrow{\$0.0021} & \dimrow{9.8} & \dimrow{1.95} & \dimrow{$-9.5$ [$-12.1$, $-7.0$]} \\
\dimrow{\jev{}} & \dimrow{--} & \dimrow{49.9} & \dimrow{\$0.0005} & \dimrow{2.1} & \dimrow{0.23} & \dimrow{$-13.4$ [$-16.8$, $-9.7$]} \\
\midrule
\multicolumn{7}{@{}l}{\emph{Cascade}} \\
\dimrow{\jev{}$\rightarrow$\gpt{}} & \dimrow{medium} & \dimrow{59.1} & \dimrow{\$0.0161} & \dimrow{74.0} & \dimrow{--} & \dimrow{$-4.2$ [$-6.5$, $-2.1$]} \\
\midrule
\multicolumn{7}{@{}l}{\emph{\method{} (decomposer: \gpt{}, verifier: \jev{})}} \\
\rowcolor{ourrow} \method{} & medium & 63.0 & \$0.0111 & 51.2 & 5.11 & $-0.3$ [$-2.9$, $+2.3$] \\
\rowcolor{ourrow} \method{} & low & \textbf{64.1} & \$0.0075 & 34.3 & 3.77 & $+0.8$ [$-1.6$, $+3.2$] \\
\rowcolor{ourrow} \method{} & none & 63.1 & \$0.0053 & 24.6 & 2.82 & $-0.2$ [$-2.8$, $+2.6$] \\
\bottomrule
\end{tabular}
\caption{Full results on the NLP reviews of \bench{} (6,466 comments, 308 factual errors). Reasoning refers to the \gpt{} or \mini{} call (for \method{}, the decomposer). Cost (\%) and $\Delta$, the paired AUROC difference, are relative to the reference. The cascade escalates 70\% of comments. Bold marks the best AUROC.}
\label{tab:main-full}
\end{table*}

\paragraph{Variant for the natural-science reviews.}
On the natural-science reviews, the decomposer labels each premise as \texttt{paper}, \texttt{knowledge}, or \texttt{opinion}.
Only \texttt{paper} premises are verified, or all premises when none is labeled \texttt{paper}.

\begin{nolinenumbers}
\begin{promptbox}{Decomposition, natural-science reviews, system prompt}
You reconstruct ONE peer-review item as a short argument, so that each step can be fact-checked separately.

List the premises the reviewer asserts or directly presupposes, and give each premise a type:
- "paper": a claim about what the paper states, reports, shows, cites or omits.
- "knowledge": a factual claim about the world or about other work that does not depend on this paper (for example what a standard method does, or what another paper introduced).
- "opinion": an evaluation, norm or recommendation (for example "X is limited", "Y is unconvincing", "the paper should do Z", "a good paper would ...").

Rules:
- Each premise is ONE self-contained declarative sentence. Say "the paper" and name the method, dataset, table, figure, section or equation it is about.
- Make "paper" and "knowledge" premises concrete and checkable. Do not use vague or evaluative words in them (little, limited, clearly, severe, sufficient, convincing, well). Put such evaluations in "opinion" premises.
- For an omission, state exactly what is missing (for example "The paper does not report results on dataset D"), not how much is missing.
- Keep each premise as close to the reviewer's own claim as possible. Do not add facts the reviewer did not assert or imply. Do not describe the paper beyond what the reviewer says. Do not judge whether a premise is true.
- If the item is a question, write only what the question presupposes about the paper.
- Use 1 to 5 premises in total.

Return ONLY JSON: {"premises": [{"text": "...", "type": "paper"}, ...], "conclusion": "..."}
\end{promptbox}
\end{nolinenumbers}

\paragraph{LLM judge prompt.}
Holistic \gpt{}, holistic \mini{}, and the LLM verifiers on paper-grounded premises use the following system prompt.

\begin{nolinenumbers}
\begin{promptbox}{LLM judge, system prompt}
You judge whether statements are true according to a given evidence text from a scientific paper. Judge each statement only from the evidence text. For each statement, give the probability (integer 0-100) that it is true according to the evidence. Return ONLY JSON: {"p": {"<id>": <int>, ...}}.
\end{promptbox}
\end{nolinenumbers}

For knowledge premises, the LLM verifiers use a neutral prompt that allows general knowledge.

\subsection{Decision-Model Questions}
\label{app:statements}

\paragraph{Questions per step.}
Table~\ref{tab:questions} lists the questions for \jev{}.
On the natural-science reviews, the contradiction question replaces the paper-grounded question, and we use one minus its probability.

\paragraph{Phrasings of the verification question.}
Table~\ref{tab:swarm-q} lists the 16 phrasings of \S\ref{sec:swarm}, which go into a single \jev{} call per premise.
Paper-grounded premises use the left column and knowledge premises the right column.
The random subsets of Figure~\ref{fig:swarm} take the $i$-th phrasings of both columns together.

\subsection{Paper Input, Cost, and Latency}
\label{app:setup}

\paragraph{Paper input.}
On the NLP reviews, all LLM systems see the same paper as \jev{}, truncated to 24k tokens.
On the natural-science reviews, holistic \gpt{} and \mini{} read the full paper with the judgment prompt of \prb{}.

\paragraph{Cost.}
Cost converts the token usage of each call into dollars at public list prices.
\gpt{} costs \$5.00 per million input tokens, \$0.50 per million cached input tokens, and \$30.00 per million output tokens, with reasoning tokens billed as output.
\mini{} costs \$0.75, \$0.075, and \$4.50 for the same three categories.
For \jev{} we use the cost reported by the API.
\kev{} runs on our own GPUs, so we report its GPU time in addition to the API cost of the decomposer.
We report cost per comment and as a percentage of the reference.

\paragraph{Latency.}
Latency is the mean wall-clock time to score one comment.
We measure it from the recorded API calls, with the premises of one comment checked in parallel.
On the natural-science reviews, we average over 300 random comments.

\FloatBarrier
\section{Additional Main Results}
\label{app:results}

This section gives the full numbers behind \S\ref{sec:results}.

\subsection{Full Results on the NLP Reviews}
\label{app:main-full}

Table~\ref{tab:main-full} extends the NLP part of Table~\ref{tab:main} with the absolute cost and latency per comment and the 95\% interval of each AUROC difference.

\subsection{Subsets}
\label{app:subsets}

Table~\ref{tab:subsets} repeats the main comparisons on human-written reviews, AI-written reviews, and the evaluation set without the pilot sample.
The pilot sample is a stratified random sample of 200 comments, 100 of them factual errors, that we scored before the full evaluation, and 194 of these comments remain after short comments are removed.
It covers \method{} with \jev{}, with \kev{} as the decision model, and with \mini{} as the verifier, as well as a comparison at equal reasoning effort.

\begin{table*}[t]
\centering
\small
\setlength{\tabcolsep}{3.2pt}
\begin{tabular}{@{}lcccccc@{}}
\toprule
 & & & \multicolumn{2}{c}{AUROC} & \multicolumn{2}{c}{$\Delta$ AUROC [95\% CI]} \\
\cmidrule(lr){4-5}\cmidrule(l){6-7}
Subset & $n$ & Errors & Holistic \gpt{} & \method{} & \method{} $-$ holistic & Verifier \\
\midrule
\multicolumn{7}{@{}l}{\emph{\method{} with \jev{}; verifier $\Delta$: \jev{} $-$ \gpt{} verifier}} \\
All & 6,466 & 308 & 63.2 & 63.0 & $-0.3$ [$-2.9$, $+2.3$] & $+0.3$ [$-1.4$, $+1.9$] \\
Human-written & 6,061 & 278 & 63.8 & 63.5 & $-0.2$ [$-3.1$, $+2.4$] & $+0.5$ [$-1.3$, $+2.3$] \\
AI-written & 405 & 30 & 64.9 & 63.1 & $-1.7$ [$-13.1$, $+8.5$] & $-1.6$ [$-7.9$, $+4.0$] \\
Without pilot sample & 6,272 & 208 & 62.1 & 63.3 & $+1.2$ [$-2.2$, $+4.3$] & $+1.1$ [$-0.9$, $+3.0$] \\
\midrule
\multicolumn{7}{@{}l}{\emph{\method{} with \kev{}; verifier $\Delta$: \kev{} $-$ \jev{}}} \\
All & 6,466 & 308 & 63.2 & 63.0 & $-0.3$ [$-3.1$, $+2.6$] & $+0.0$ [$-2.0$, $+1.9$] \\
Human-written & 6,061 & 278 & 63.8 & 63.2 & $-0.5$ [$-3.6$, $+2.6$] & $-0.3$ [$-2.6$, $+1.8$] \\
AI-written & 405 & 30 & 64.9 & 67.8 & $+2.9$ [$-8.6$, $+12.9$] & $+4.6$ [$-1.1$, $+11.5$] \\
Without pilot sample & 6,272 & 208 & 62.1 & 63.5 & $+1.4$ [$-1.7$, $+4.6$] & $+0.2$ [$-2.1$, $+2.5$] \\
\midrule
\multicolumn{7}{@{}l}{\emph{\method{} with the \mini{} verifier; verifier $\Delta$: \mini{} $-$ \jev{}}} \\
All & 6,466 & 308 & 63.2 & 62.4 & $-0.8$ [$-3.5$, $+2.2$] & $-0.5$ [$-2.4$, $+1.6$] \\
Human-written & 6,061 & 278 & 63.8 & 62.4 & $-1.3$ [$-4.1$, $+1.8$] & $-1.1$ [$-3.0$, $+0.9$] \\
AI-written & 405 & 30 & 64.9 & 67.4 & $+2.5$ [$-7.8$, $+12.3$] & $+4.3$ [$-1.5$, $+9.7$] \\
Without pilot sample & 6,272 & 208 & 62.1 & 62.5 & $+0.4$ [$-2.7$, $+3.8$] & $-0.7$ [$-3.0$, $+1.8$] \\
\midrule
\multicolumn{7}{@{}l}{\emph{Equal reasoning effort: \method{} with the low decomposer vs.\ holistic \gpt{} at low}} \\
All & 6,466 & 308 & 61.2 & 64.1 & $+2.9$ [$+0.3$, $+5.3$] & -- \\
Human-written & 6,061 & 278 & -- & -- & $+3.2$ [$+0.8$, $+5.7$] & -- \\
AI-written & 405 & 30 & -- & -- & $+1.9$ [$-12.9$, $+14.7$] & -- \\
Without pilot sample & 6,272 & 208 & -- & -- & $+3.3$ [$-0.0$, $+6.2$] & -- \\
\bottomrule
\end{tabular}
\caption{Subset results on the NLP reviews. \method{} uses the medium-reasoning decomposer in the upper three blocks. Verifier $\Delta$ compares two verifiers on the same premises. Intervals use a paired bootstrap over papers.}
\label{tab:subsets}
\end{table*}

\subsection{Latency}
\label{app:latency}

Table~\ref{tab:latency} gives the latency measurements of \S\ref{sec:time}.
The upper block splits the recorded time of \method{} by step and compares verifiers, and the lower block gives the controlled run.

\begin{table}[!ht]
\begin{minipage}[c]{0.52\textwidth}
\centering
\footnotesize
\setlength{\tabcolsep}{2pt}
\begin{adjustbox}{max width=\linewidth}
\begin{tabular}{@{}lcccc@{}}
\toprule
System & Mean & Median & p90 & Ratio [95\% CI] \\
\midrule
\multicolumn{5}{@{}l}{\emph{Recorded API calls}} \\
\method{} (none) & 2.82 & -- & -- & -- \\
\quad decomposer & 2.33 & -- & -- & -- \\
\quad \jev{} & 0.49 & -- & -- & -- \\
\multicolumn{5}{@{}l}{Verifier on the medium-decomposer premises} \\
\quad \jev{} & 0.44 & -- & -- & -- \\
\quad \mini{} & 2.21 & -- & -- & -- \\
\quad \gpt{} & 7.38 & -- & -- & -- \\
\midrule
\multicolumn{5}{@{}l}{\emph{Controlled run, 113 comments}} \\
Holistic \gpt{} & 7.50 & 6.33 & 9.17 & 1.00 \\
\rowcolor{ourrow} \method{} (none) & 3.18 & 2.69 & 3.32 & 0.42 [0.30, 0.64] \\
\rowcolor{ourrow} \method{} (medium) & 5.62 & 4.82 & 8.38 & 0.75 [0.58, 0.96] \\
\dimrow{Batched, review} & \dimrow{2.85} & \dimrow{2.50} & \dimrow{3.65} & \dimrow{0.38 [0.28, 0.48]} \\
\dimrow{Batched, comment} & \dimrow{0.17} & \dimrow{0.13} & \dimrow{0.27} & \dimrow{0.02 [0.02, 0.03]} \\
\bottomrule
\end{tabular}
\end{adjustbox}
\end{minipage}\hfill
\begin{minipage}[c]{0.44\textwidth}
\caption{Latency in seconds on the NLP reviews. Parentheses give the reasoning of the decomposer. In the upper block, the \jev{} rows include routing, and the \mini{} and \gpt{} rows check the same premises instead. The ratio compares mean latency with holistic \gpt{}. The batched rows give batched holistic \gpt{} without reasoning per review and per comment.}
\label{tab:latency}
\end{minipage}
\end{table}

\subsection{Robustness}
\label{app:robust}

Table~\ref{tab:robust} gives the numbers behind Figure~\ref{fig:robust}.

\begin{table}[!ht]
\begin{minipage}[c]{0.63\textwidth}
\centering
\footnotesize
\setlength{\tabcolsep}{2.5pt}
\begin{adjustbox}{max width=\linewidth}
\begin{tabular}{@{}lcccc@{}}
\toprule
 & \multicolumn{2}{c}{AUROC} & & \\
\cmidrule(lr){2-3}
Setting & \begin{tabular}[c]{@{}c@{}}Holistic\\\gpt{}\end{tabular} & \methodhd{} & $\Delta$ [95\% CI] & \begin{tabular}[c]{@{}c@{}}Cost\\(\%)\end{tabular} \\
\midrule
\multicolumn{5}{@{}l}{\emph{Open-weight decision model \kev{}}} \\
Medium decomposer & 63.2 & 63.0 & $-0.3$ [$-3.1$, $+2.6$] & 44.1\,+\,GPU \\
No-reasoning decomposer & 63.2 & 63.8 & $+0.6$ [$-2.6$, $+3.6$] & 17.8\,+\,GPU \\
\midrule
\multicolumn{5}{@{}l}{\emph{Comment length (words)}} \\
Short ($\le$13) & 61.3 & 59.8 & $-1.6$ [$-7.6$, $+4.5$] & 20.9 \\
Medium (14--21) & 67.1 & 64.2 & $-2.9$ [$-7.5$, $+1.9$] & 24.7 \\
Long ($>$21) & 65.6 & 62.7 & $-2.9$ [$-7.2$, $+1.7$] & 28.0 \\
Combined with length & 71.9 & 70.2 & $-1.7$ [$-3.3$, $+0.1$] & 24.6 \\
\bottomrule
\end{tabular}
\end{adjustbox}
\end{minipage}\hfill
\begin{minipage}[c]{0.34\textwidth}
\caption{Holistic \gpt{} (medium reasoning) and \method{} on the NLP reviews in further settings. $\Delta$ is the paired difference. The length groups split the comments at the tertiles of comment length, and the last row averages the percentile ranks of each score and of comment length. Cost is that of \method{} relative to holistic \gpt{}, plus GPU time for \kev{}, which has no API price.}
\label{tab:robust}
\end{minipage}

\vspace{14pt}
\begin{minipage}[c]{0.63\textwidth}
\centering
\footnotesize
\setlength{\tabcolsep}{3pt}
\begin{adjustbox}{max width=\linewidth}
\begin{tabular}{@{}lccc@{}}
\toprule
System & AUROC & Cost (\%) & $\Delta$ [95\% CI] \\
\midrule
\multicolumn{4}{@{}l}{\emph{Machine learning reviews (216 comments, 46 errors)}} \\
Holistic \gpt{} & 66.9 & 100.0 & -- \\
\dimrow{Holistic \jev{}} & \dimrow{61.2} & \dimrow{1.6} & \dimrow{$-5.7$ [$-14.7$, $+2.0$]} \\
\dimrow{\method{}, NLP prompt} & \dimrow{60.8} & \dimrow{49.5} & \dimrow{$-6.0$ [$-16.2$, $+3.1$]} \\
\rowcolor{ourrow} \method{}, ML prompt & 68.7 & 46.6 & $+1.8$ [$-9.4$, $+11.7$] \\
\midrule
\multicolumn{4}{@{}l}{\emph{Natural-science reviews (2,585 comments, 65 errors)}} \\
Holistic \gpt{} & 67.8 & 100.0 & -- \\
\dimrow{\method{}, supports question} & \dimrow{63.3} & \dimrow{--} & \dimrow{$-4.4$} \\
\rowcolor{ourrow} \method{}, contradicts question & 67.5 & 47.4 & $-0.3$ [$-8.9$, $+7.6$] \\
\quad human-written (40 errors) & 66.8 & -- & $+1.2$ [$-10.3$, $+11.7$] \\
\quad AI-written (25 errors) & 69.1 & -- & $-2.9$ [$-16.0$, $+10.6$] \\
\bottomrule
\end{tabular}
\end{adjustbox}
\end{minipage}\hfill
\begin{minipage}[c]{0.34\textwidth}
\caption{Results on the machine learning and natural-science reviews. $\Delta$ is the paired AUROC difference to holistic \gpt{} on the same comments, and cost is relative to it. Both natural-science \method{} rows use the natural-science decomposition prompt.}
\label{tab:other}
\end{minipage}
\end{table}

\subsection{Machine Learning and Natural-Science Reviews}
\label{app:other-parts}

Table~\ref{tab:other} gives the results on the machine learning and natural-science reviews and compares decomposition prompts and verification questions on them.
On the machine learning reviews, the machine learning prompt yields a shorter premise list. It leaves out premises that the reviewer did not state and so reduces false alarms on correct comments.
On the natural-science reviews, the choice of verification question matters most.
Asking whether the paper supports a premise treats details that the paper does not mention as false.
Asking whether the paper contradicts the premise avoids this.

\section{Additional Analysis}
\label{app:analysis}

This section gives the full numbers behind \S\ref{sec:analysis}.

\subsection{Capability Map of the Decision Model}
\label{app:capmap}

\begin{figure}[!ht]
\centering
\begin{minipage}[c]{0.40\textwidth}
\centering
\includegraphics[width=\linewidth]{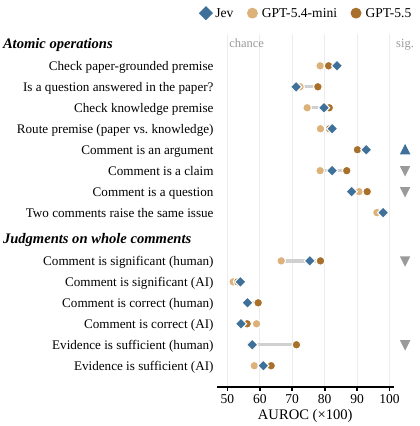}
\end{minipage}\hfill
\begin{minipage}[c]{0.55\textwidth}
\caption{AUROC of \jev{}, \mini{}, and \gpt{} on each verification operation. Triangles mark operations on which \jev{} is significantly better ($\blacktriangle$) or worse ($\blacktriangledown$) than \gpt{}.}
\label{fig:capmap}
\end{minipage}
\end{figure}

Table~\ref{tab:capmap-full} measures \jev{} on individual operations of review verification, and Figure~\ref{fig:capmap} plots the results.
For each operation, \jev{}, \mini{} (low reasoning), and \gpt{} (medium reasoning) receive the same input and questions, and the two LLMs return a probability as an integer from 0 to 100.
Premise checking and routing use \rs{} \citep{ryu2025reviewscore}: 452 human-reconstructed premises about the paper, 33 of them false, and the grounding source annotators gave for 972 premises.
Comment-type classification uses \rs{} comments.
Duplicate detection and all judgments use \prb{} \citep{kim2026peerreviewbench}.

\begin{table*}[t]
\centering
\small
\setlength{\tabcolsep}{5pt}
\begin{tabular}{@{}lrrcccc@{}}
\toprule
 & & & \multicolumn{3}{c}{AUROC} & \\
\cmidrule(lr){4-6}
Operation & $n$ & Neg. & \jev{} & \mini{} & \gpt{} & \jev{} $-$ \gpt{} [95\% CI] \\
\midrule
\multicolumn{7}{@{}l}{\emph{Atomic operations}} \\
Check paper-grounded premise & 452 & 33 & \textbf{83.8} & 78.6 & 81.2 & $+2.6$ [$-1.0$, $+6.0$] \\
Is a question answered in the paper? & 117 & 83 & 71.2 & 72.4 & \textbf{77.9} & $-6.7$ [$-14.8$, $+1.2$] \\
Check knowledge premise & 481 & 21 & 79.8 & 74.6 & \textbf{81.4} & $-1.6$ [$-10.8$, $+9.4$] \\
Route premise (paper vs.\ knowledge) & 972 & 499 & \textbf{82.3} & 78.7 & 81.5 & $+0.8$ [$-2.1$, $+4.1$] \\
Comment is an argument & 467 & 178 & \textbf{92.8} & 90.2 & 90.1 & $+2.7$ [$+0.2$, $+5.0$] \\
Comment is a claim & 467 & 387 & 82.3 & 78.6 & \textbf{86.8} & $-4.5$ [$-8.2$, $-0.7$] \\
Comment is a question & 467 & 369 & 88.3 & 90.6 & \textbf{93.1} & $-4.8$ [$-7.1$, $-2.7$] \\
Two comments raise the same issue & 164 & 94 & 98.0 & 96.1 & \textbf{98.4} & $-0.5$ [$-1.6$, $+0.5$] \\
\midrule
\multicolumn{7}{@{}l}{\emph{Judgments on whole comments}} \\
Comment is significant (human) & 1,748 & 908 & 75.4 & 66.6 & \textbf{78.7} & $-3.3$ [$-5.8$, $-0.9$] \\
Comment is significant (AI) & 869 & 299 & \textbf{54.0} & 51.8 & 53.2 & $+0.8$ [$-4.1$, $+5.7$] \\
Comment is correct (human) & 1,930 & 182 & 56.2 & 56.2 & \textbf{59.5} & $-3.2$ [$-7.8$, $+1.1$] \\
Comment is correct (AI) & 1,029 & 160 & 54.2 & \textbf{59.0} & 56.1 & $-1.9$ [$-6.8$, $+2.3$] \\
Evidence is sufficient (human) & 1,360 & 135 & 57.7 & 57.2 & \textbf{71.3} & $-13.7$ [$-19.5$, $-8.1$] \\
Evidence is sufficient (AI) & 793 & 38 & 61.1 & 58.3 & \textbf{63.5} & $-2.4$ [$-11.2$, $+6.7$] \\
\bottomrule
\end{tabular}
\caption{AUROC of three models on verification operations with identical inputs. $n$ is the number of items and Neg.\ the number of negative items. Bold marks the best model in each row.}
\label{tab:capmap-full}
\end{table*}

On the operations that \method{} uses, \jev{} is as accurate as \gpt{}: it checks a single premise against the full paper at least as well, and it routes premises equally well.
On classifying the type of a comment, it is significantly better at recognizing arguments but significantly worse at recognizing claims and questions.
On judgments about a whole comment, \jev{} falls behind, and on whether a comment is correct, all models are close to chance.
A premise check that reads the full paper costs \$0.00044 with \jev{}, and its median latency is 216\,ms compared with 5.1\,s for \gpt{}.

\subsection{Verifiers}
\label{app:checkers}

Table~\ref{tab:checker} gives the AUROC of each model as a holistic judge and as the verifier of \method{}, together with the paired differences and the latency, and Table~\ref{tab:minicheck} adds MiniCheck.
As a holistic judge, \kev{} differs from holistic \gpt{} by $-8.9$ [$-11.8$, $-6.0$].

\begin{table*}[t]
\centering
\small
\setlength{\tabcolsep}{4.5pt}
\begin{tabular}{@{}lcccccccc@{}}
\toprule
 & \multicolumn{2}{c}{AUROC} & \multicolumn{2}{c}{Verifier $\Delta$ [95\% CI]} & \multicolumn{2}{c}{Cost / comment} & \multicolumn{2}{c}{Latency (s)} \\
\cmidrule(lr){2-3}\cmidrule(lr){4-5}\cmidrule(lr){6-7}\cmidrule(l){8-9}
Model & Holistic & Verifier & to reference & to \jev{} verifier & \$ & \% & Total & Verify \\
\midrule
\gpt{} & \textbf{63.2} & 62.7 & $-0.6$ [$-3.1$, $+2.1$] & $-0.3$ [$-1.9$, $+1.4$] & 0.0678 & 312.7 & 12.06 & 7.38 \\
\mini{} & 53.7 & 62.4 & $-0.8$ [$-3.5$, $+2.2$] & $-0.5$ [$-2.4$, $+1.6$] & 0.0156 & 71.9 & 6.90 & 2.21 \\
\rowcolor{ourrow} \jev{} & 49.9 & \textbf{63.0} & $-0.3$ [$-2.9$, $+2.3$] & -- & 0.0111 & 51.2 & 5.11 & 0.44 \\
\rowcolor{ourrow} \kev{} & 54.3 & \textbf{63.0} & $-0.3$ [$-3.1$, $+2.6$] & $+0.0$ [$-2.0$, $+1.9$] & 0.0096 + GPU & 44.1 & -- & -- \\
\bottomrule
\end{tabular}
\caption{AUROC on the NLP reviews of each model as a holistic judge and as the verifier of \method{}. All verifiers receive the same premises from the medium-reasoning decomposer and the same routing. The reference is holistic \gpt{} at medium reasoning, and cost (\%) is relative to it. Cost and total latency cover the whole pipeline, and verification latency covers only the verification step. Holistic \mini{} uses low reasoning.}
\label{tab:checker}
\end{table*}

\paragraph{Off-the-shelf fact-checking models.}
\label{app:minicheck}

We replace \jev{} with MiniCheck \citep{tang2024minicheck}, a family of small models trained to test whether a document supports a claim.
The main variant is Bespoke-MiniCheck-7B (MiniCheck-7B), and we also run MiniCheck-Flan-T5-Large (MiniCheck-T5).
Both read the same truncated paper as \jev{} and use the default chunking and aggregation of the MiniCheck library.
The premises are those of \method{} with the medium-reasoning and the no-reasoning decomposer, and the comment score is again the minimum over premises.
MiniCheck takes no question, so every premise is checked against the paper.
In the hybrid variant, \jev{} checks the knowledge premises instead.
GPU time is the wall-clock time of each stage on eight H200 GPUs divided by the number of comments, as for \kev{}.
Table~\ref{tab:minicheck} gives the results.

\begin{table*}[t]
\centering
\small
\setlength{\tabcolsep}{3.5pt}
\begin{tabular}{@{}llccccc@{}}
\toprule
 & & & API cost & GPU & \multicolumn{2}{c}{$\Delta$ [95\% CI]} \\
\cmidrule(l){6-7}
System & Reasoning & AUROC & (\%) & (s) & to reference & to \jev{} verifier \\
\midrule
\multicolumn{7}{@{}l}{\emph{Holistic judgment}} \\
\gpt{} (ref.) & medium & 63.2 & 100.0 & -- & -- & -- \\
MiniCheck-7B & -- & 59.5 & 0 & 0.03 & $-3.7$ [$-6.9$, $-0.3$] & -- \\
\dimrow{MiniCheck-T5} & \dimrow{--} & \dimrow{56.6} & \dimrow{0} & \dimrow{0.69} & \dimrow{$-6.7$ [$-9.8$, $-3.4$]} & \dimrow{--} \\
\midrule
\multicolumn{7}{@{}l}{\emph{\method{} with different verifiers}} \\
\rowcolor{ourrow} \jev{} & medium & \textbf{63.0} & 51.2 & -- & $-0.3$ [$-2.9$, $+2.3$] & -- \\
\kev{} & medium & 63.0 & 44.1 & 3.7 & $-0.3$ [$-3.1$, $+2.6$] & $+0.0$ [$-2.0$, $+1.9$] \\
MiniCheck-7B & medium & 54.0 & 44.1 & 0.06 & $-9.2$ [$-12.9$, $-5.6$] & $-8.9$ [$-12.3$, $-5.5$] \\
MiniCheck-7B & none & 53.9 & 17.8 & 0.06 & $-9.3$ [$-13.2$, $-5.4$] & $-9.1$ [$-12.3$, $-6.0$] \\
\dimrow{MiniCheck-T5} & \dimrow{medium} & \dimrow{52.7} & \dimrow{44.1} & \dimrow{2.18} & \dimrow{$-10.5$ [$-14.2$, $-6.7$]} & \dimrow{$-10.3$ [$-13.8$, $-6.6$]} \\
\dimrow{MiniCheck-T5} & \dimrow{none} & \dimrow{53.6} & \dimrow{17.8} & \dimrow{2.04} & \dimrow{$-9.7$ [$-13.1$, $-6.0$]} & \dimrow{$-9.5$ [$-12.5$, $-6.2$]} \\
\dimrow{MiniCheck-7B, \jev{} for knowledge} & \dimrow{medium} & \dimrow{56.8} & \dimrow{47.3} & \dimrow{--} & \dimrow{--} & \dimrow{$-6.2$ [$-8.9$, $-3.4$]} \\
\bottomrule
\end{tabular}
\caption{Off-the-shelf fact-checking models on the NLP reviews (6,466 comments, 308 factual errors). Reasoning refers to the \gpt{} call (for \method{}, the decomposer). API cost is relative to the reference, and for \kev{} and MiniCheck it covers only the decomposer. GPU is H200 GPU seconds per comment, an upper bound that includes idle time. The last column compares with \method{} using \jev{} on the same premises.}
\label{tab:minicheck}
\end{table*}

\begin{table}[t]
\begin{minipage}[t]{0.52\textwidth}
\vspace{0pt}
\centering
\footnotesize
\setlength{\tabcolsep}{3pt}
\begin{adjustbox}{max width=\linewidth}
\begin{tabular}{@{}lccccc@{}}
\toprule
 & & \multicolumn{4}{c}{Share of cost (\%)} \\
\cmidrule(l){3-6}
Call & Cost & Input & Cached & Reasoning & Output \\
\midrule
\multicolumn{6}{@{}l}{\emph{\gpt{} calls by token type}} \\
Holistic & & & & & \\
\quad medium & \$0.0217 & 27 & 20 & 49 & 4 \\
\dimrow{\quad low} & \dimrow{\$0.0165} & \dimrow{34} & \dimrow{27} & \dimrow{34} & \dimrow{5} \\
\dimrow{\quad none} & \dimrow{\$0.0108} & \dimrow{53} & \dimrow{41} & \dimrow{0} & \dimrow{6} \\
Decomposer & & & & & \\
\quad medium & \$0.0096 & 13 & 0 & 57 & 30 \\
\dimrow{\quad low} & \dimrow{\$0.0060} & \dimrow{22} & \dimrow{0} & \dimrow{32} & \dimrow{46} \\
\dimrow{\quad none} & \dimrow{\$0.0039} & \dimrow{33} & \dimrow{0} & \dimrow{0} & \dimrow{67} \\
\midrule
\multicolumn{3}{@{}l}{\emph{\method{} (medium) by stage}} & \multicolumn{3}{c}{\emph{Share (\%)}} \\
Decomposer & \$0.0096 & & \multicolumn{3}{c}{86} \\
\jev{} & \$0.0015 & & \multicolumn{3}{c}{14} \\
\dimrow{\mini{}} & \dimrow{\$0.0060} & & \multicolumn{3}{c}{\dimrow{--}} \\
\dimrow{\gpt{}} & \dimrow{\$0.058} & & \multicolumn{3}{c}{\dimrow{--}} \\
\midrule
\multicolumn{6}{@{}l}{\emph{Saving of \method{} (none) vs.\ holistic \gpt{}}} \\
\quad medium & 75.4\% & & & & \\
\quad low & 68\% & & & & \\
\bottomrule
\end{tabular}
\end{adjustbox}
\caption{Per-comment cost on the NLP reviews. Shares split the cost of each \gpt{} call by token type and the cost of \method{} by stage. The \jev{} row includes routing, and the dimmed \mini{} and \gpt{} rows give the approximate cost of checking with these models instead.}
\label{tab:cost}
\end{minipage}\hfill
\begin{minipage}[t]{0.44\textwidth}
\vspace{0pt}
\centering
\footnotesize
\setlength{\tabcolsep}{2.7pt}
\begin{tabular}{@{}lccc@{}}
\toprule
 & \multicolumn{2}{c}{Cost / comment} & \\
\cmidrule(lr){2-3}
Paper length (tokens) & \begin{tabular}[c]{@{}c@{}}Holistic\\\gpt{}\end{tabular} & \methodhd{} & Cost (\%) \\
\midrule
4.9k--8.1k & \$0.0200 & \$0.0107 & 53.3 \\
8.1k--9.6k & \$0.0211 & \$0.0111 & 52.4 \\
9.7k--24k & \$0.0240 & \$0.0116 & 48.5 \\
\bottomrule
\end{tabular}
\caption{Cost per comment on the NLP reviews by paper length. Holistic \gpt{} and the decomposer use medium reasoning. Each length group holds one third of the comments, and the median paper length is about 8.5k tokens. Cost (\%) is relative to holistic \gpt{}.}
\label{tab:paperlen}
\end{minipage}
\end{table}

MiniCheck-7B lowers \method{} from 63.0 to 54.0, and MiniCheck-T5 behaves the same.
The cause is false alarms on correct premises.
Among premises that do not claim an omission, MiniCheck-7B gives 25.5\% a probability below 0.3, compared with 11.2\% for \jev{}.
A comment has 3.3 premises on average, so with the minimum a single false alarm marks the whole comment as wrong.
Using \jev{} for the knowledge premises recovers only 2.8, so most of the gap lies on premises that are checked against the paper.

Holistic MiniCheck-7B, which scores the whole comment against the paper, reaches 59.5 at no API cost.
\method{} with a single question is $+3.4$ [$-0.0$, $+6.9$] above it.
With the 16 phrasings per premise of \S\ref{sec:swarm}, \method{} is $+6.1$ [$+3.0$, $+9.0$] above it.

\subsection{Cost Breakdown and Paper Length}
\label{app:cost}

Table~\ref{tab:cost} gives the per-comment costs on the NLP reviews behind Figure~\ref{fig:cost}.
The first block splits the cost of holistic \gpt{} and of the \gpt{} decomposer by token type.
At medium reasoning, holistic judgment reads on average 9,991 input tokens per comment, of which 8,831 hit the prompt cache, and produces 383 output tokens, of which 353 are reasoning tokens.
The decomposer reads only the comment.

The second block gives the cost of the two stages of \method{}, the decomposer and \jev{}.
\jev{} costs \$0.0015 per comment with the low- and no-reasoning decomposers as well.
The \mini{} and \gpt{} rows give the cost of the verifiers that replace \jev{} in \S\ref{sec:checker}.
The last block gives the saving of \method{} with the no-reasoning decomposer over holistic \gpt{} at medium and at low reasoning.

Table~\ref{tab:paperlen} gives the cost per comment by paper length behind Figure~\ref{fig:cost}.

\subsection{Other Decision Models and Other LLMs}
\label{app:models}

Table~\ref{tab:models} gives the full results of \S\ref{sec:models}.
The decision models receive the same requests as \jev{} through the same API, and all other settings follow \S\ref{sec:setup}.
The LLMs receive the decomposition and holistic prompts of \gpt{} unchanged.
GPT-5.6 Sol and GPT-5.6 Luna use medium reasoning, like \gpt{}.
The other LLMs do not reason and use temperature 0, except Nemotron Super 3, which uses its default sampling because it produces malformed output at temperature 0.
A holistic answer from which no probability can be read is scored 0.5, which affects 2.1\% of the comments for Nemotron Super 3, 0.6\% for Ministral 3 14B, and fewer than 0.1\% for the other LLMs.
API costs use public prices, and \kev{} runs on our own GPUs, so its checks add no API cost.

\begin{table*}[t]
\centering
\small
\setlength{\tabcolsep}{4pt}
\begin{adjustbox}{max width=\textwidth}
\begin{tabular}{@{}lccccccc@{}}
\toprule
 & \multicolumn{2}{c}{AUROC} & \multicolumn{2}{c}{Cost (\%)} & \multicolumn{3}{c}{\method{} minus [95\% CI]} \\
\cmidrule(lr){2-3}\cmidrule(lr){4-5}\cmidrule(l){6-8}
Model & Holistic & \method{} & own & \gpt{} & own holistic & holistic \gpt{} & default \\
\midrule
\multicolumn{8}{@{}l}{\emph{Other decision models}, with the medium-reasoning \gpt{} decomposer (6,394 comments, 304 errors)} \\
\dimrow{\jev{} (default)} & \dimrow{49.9} & \dimrow{63.1} & & \dimrow{51.3} & & \dimrow{$-0.2$} & \dimrow{--} \\
Microsoft Decision-1 & 53.6 & 64.3 & & 50.9 & & $+1.0$ [$-2.2$, $+4.3$]$^\dagger$ & $+1.2$ [$-0.9$, $+3.5$]$^\dagger$ \\
Perplexity Decider & 57.2 & 62.6 & & 47.5 & & $-0.7$ [$-3.6$, $+2.5$] & $-0.5$ [$-2.5$, $+1.6$]$^\dagger$ \\
\midrule
\multicolumn{8}{@{}l}{\emph{Other LLMs} as decomposer and holistic judge, with \kev{} as router and verifier (6,466 comments, 308 errors)} \\
\dimrow{\gpt{} (default)} & \dimrow{63.2} & \dimrow{63.0} & \dimrow{44.1} & \dimrow{44.1} & \dimrow{--} & \dimrow{$-0.3$} & \dimrow{--} \\
Nemotron Super 3 & 43.9 & 65.4 & 6.0 & 0.4 & $+21.6$ [$+17.3$, $+25.6$]$^\dagger$ & $+2.2$ [$-0.8$, $+4.9$]$^\dagger$ & $+2.4$ [$-0.5$, $+5.4$]$^\dagger$ \\
GPT-5.6 Sol & 61.0 & 64.1 & 9.6 & 9.4 & $+3.1$ [$-0.4$, $+6.5$]$^\dagger$ & $+0.9$ [$-2.2$, $+3.8$]$^\dagger$ & $+1.1$ [$-1.2$, $+3.4$]$^\dagger$ \\
GPT-5.6 Luna & 52.7 & 63.3 & 17.1 & 1.7 & $+10.7$ [$+7.1$, $+13.7$]$^\dagger$ & $+0.1$ [$-2.8$, $+2.7$]$^\dagger$ & $+0.4$ [$-2.0$, $+2.7$]$^\dagger$ \\
Claude Haiku 5.5 & 47.4 & 61.2 & 8.8 & 0.7 & $+13.9$ [$+9.3$, $+18.4$]$^\dagger$ & $-2.0$ [$-5.5$, $+1.4$] & $-1.8$ [$-4.1$, $+0.6$] \\
Ministral 3 14B & 49.9 & 61.1 & 3.9 & 0.4 & $+11.2$ [$+6.8$, $+15.2$]$^\dagger$ & $-2.2$ [$-4.9$, $+0.7$] & $-1.9$ [$-4.5$, $+0.6$] \\
Gemma 3 27B & 48.6 & 57.5 & 4.1 & 0.4 & $+8.9$ [$+4.8$, $+12.8$]$^\dagger$ & $-5.8$ [$-8.8$, $-2.7$] & $-5.5$ [$-8.1$, $-2.9$] \\
\bottomrule
\end{tabular}
\end{adjustbox}
\caption{Other decision models and other LLMs on the NLP reviews of \bench{}. The upper block leaves out one paper that Microsoft Decision-1 cannot read within its context window. Cost is the API cost of \method{} relative to the holistic judgment of \gpt{} and, in the lower block, of the LLM in the row (own). The last column compares with the default configuration of each block. Differences are in AUROC points with paired 95\% intervals from a bootstrap over papers, and $^\dagger$ marks non-inferiority with a margin of $-3$.}
\label{tab:models}
\end{table*}

\subsection{Design Choices}
\label{app:design}

Table~\ref{tab:design} gives the decomposition statistics and design choices of \S\ref{sec:design}.

Table~\ref{tab:swarm-ablation} compares the sets of \jev{} phrasings of \S\ref{sec:swarm} with the single question.

\begin{table*}[t]
\begin{minipage}[t]{0.56\textwidth}
\vspace{0pt}
\centering
\small
\setlength{\tabcolsep}{4pt}
\begin{tabular}{@{}lcc@{}}
\toprule
\multicolumn{3}{@{}l}{\emph{Decomposition on the NLP reviews}} \\
\multicolumn{2}{@{}l}{Premises per comment} & 3.28 \\
\multicolumn{2}{@{}l}{\quad routed as knowledge premises by \jev{}} & 1.42 \\
\multicolumn{2}{@{}l}{Premises in total} & 21,228 \\
\multicolumn{2}{@{}l}{Routing agreement of \kev{} with \jev{} (\%)} & 65.2 \\
\midrule
\emph{Design choices (AUROC)} & ML & Overall \\
\rowcolor{ourrow} \jev{} routing, minimum & 61.8 & 53.3 \\
\quad all premises checked against the paper & 58.8 & 50.1 \\
\quad mean instead of minimum & 59.4 & 54.7 \\
\bottomrule
\end{tabular}
\caption{Decomposition statistics and design choices of \method{}. On the NLP reviews, the decomposer uses medium reasoning. Design choices are scored on the premises of an earlier version of the NLP decomposition prompt, so ML differs from Table~\ref{tab:main}, by AUROC on the machine learning reviews (ML) and on the overall factuality rating of \rs{} (Overall), which also counts errors of judgment. Each indented row changes one choice of the first row.}
\label{tab:design}
\end{minipage}\hfill
\begin{minipage}[t]{0.40\textwidth}
\vspace{0pt}
\centering
\small
\setlength{\tabcolsep}{4pt}
\begin{tabular}{@{}lcc@{}}
\toprule
\jev{} phrasings & AUROC & $\Delta$ [95\% CI] \\
\midrule
\multicolumn{3}{@{}l}{\emph{Decomposer: \gpt{}, no reasoning}} \\
Single question & 63.1 & -- \\
8 supporting & 63.0 & $-0.0$ [$-0.4$, $+0.4$] \\
8 contradiction & 67.6 & $+4.5$ [$+2.7$, $+6.3$] \\
\rowcolor{ourrow} All 16 & 65.7 & $+2.6$ [$+1.7$, $+3.5$] \\
\midrule
\multicolumn{3}{@{}l}{\emph{Decomposer: \gpt{}, medium reasoning}} \\
Single question & 63.0 & -- \\
8 contradiction & 67.5 & -- \\
\rowcolor{ourrow} All 16 & 65.2 & $+2.2$ [$+1.3$, $+3.1$] \\
\bottomrule
\end{tabular}
\caption{AUROC of \method{} on the NLP reviews with different sets of \jev{} phrasings per premise. All sets have the same cost. $\Delta$ is the paired difference to the first row of each block.}
\label{tab:swarm-ablation}
\end{minipage}
\end{table*}

\begin{table*}[t]
\centering
\footnotesize
\setlength{\tabcolsep}{3pt}
\begin{adjustbox}{max width=\textwidth}
\begin{tabular}{@{}lrcccccccc@{}}
\toprule
 & & \multicolumn{4}{c}{Per comment} & \multicolumn{3}{c}{Batched per review} & \\
\cmidrule(lr){3-6}\cmidrule(lr){7-9}
 & & Holistic & \multicolumn{2}{c}{\method{}} & & Holistic & \methodhd{} & & Omission \\
\cmidrule(lr){4-5}
Error type & $n$ & medium & medium & $k{=}16$ & $\Delta$ [95\% CI] & none & $k{=}16$ & $\Delta$ [95\% CI] & (\%) \\
\midrule
All errors & 308 & 63.2 & 63.0 & 65.7 & $+2.4$ [$-0.3$, $+5.3$] & 67.8 & 69.0 & $+1.2$ [$-1.2$, $+4.0$] & 56.5 \\
\quad Misunderstanding & 149 & 59.5 & 61.0 & 62.7 & $+3.2$ [$-1.2$, $+7.3$] & 63.6 & 65.7 & $+2.1$ [$-1.6$, $+6.0$] & 39.2 \\
\quad Neglect & 145 & 67.8 & 65.2 & 69.9 & $+2.1$ [$-1.1$, $+5.4$] & 72.6 & 73.1 & $+0.5$ [$-3.8$, $+5.0$] & 78.5 \\
\dimrow{\quad Unstated statement} & \dimrow{8} & \dimrow{70.5} & \dimrow{78.4} & \dimrow{61.4} & \dimrow{$-9.0$ [$-20.9$, $+7.7$]} & \dimrow{74.5} & \dimrow{74.4} & \dimrow{$-0.1$ [$-12.8$, $+19.6$]} & \dimrow{12.5} \\
\dimrow{\quad Inaccurate summary} & \dimrow{6} & \dimrow{37.1} & \dimrow{35.5} & \dimrow{42.0} & \dimrow{$+4.9$ [$-13.3$, $+22.3$]} & \dimrow{48.8} & \dimrow{48.8} & \dimrow{$-0.0$ [$-15.6$, $+17.1$]} & \dimrow{16.7} \\
\midrule
\dimrow{Correct comments} & \dimrow{6,158} & \dimrow{--} & \dimrow{--} & \dimrow{--} & \dimrow{--} & \dimrow{--} & \dimrow{--} & \dimrow{--} & \dimrow{35.9} \\
\bottomrule
\end{tabular}
\end{adjustbox}
\caption{AUROC on the NLP reviews by error type. Each row compares the errors of one type with all correct comments. Holistic is \gpt{} at the given reasoning, and \method{} uses the medium-reasoning decomposer or, with $k{=}16$, the no-reasoning decomposer with 16 phrasings per premise. Each $\Delta$ compares \method{} ($k{=}16$) with the holistic column of its block. Omission is the share of comments whose lowest premise under \method{} (medium) claims an omission.}
\label{tab:error-types}
\end{table*}

\begin{table*}[t]
\centering
\small
\setlength{\tabcolsep}{6pt}
\begin{tabular}{@{}lcccc@{}}
\toprule
 & \multicolumn{2}{c}{Factual label} & \multicolumn{2}{c}{Judgment label} \\
\cmidrule(lr){2-3}\cmidrule(l){4-5}
System & AUROC & $\Delta$ [95\% CI] & AUROC & $\Delta$ [95\% CI] \\
\midrule
\multicolumn{5}{@{}l}{\emph{Natural-science reviews} (factual: 65 errors; judgment: 315 errors)} \\
Holistic \gpt{} & 67.8 & -- & 57.8 & -- \\
\rowcolor{ourrow} \method{} & 67.5 & $-0.3$ [$-8.9$, $+7.6$] & 53.5 & $-4.4$ [$-8.4$, $-0.2$] \\
\midrule
\multicolumn{5}{@{}l}{\emph{Machine learning reviews} (factual: 46 of 216 comments; judgment: 16 of 296 comments)} \\
Holistic \gpt{} & 66.9 & -- & 81.5 & -- \\
\rowcolor{ourrow} \method{} & 68.7 & $+1.8$ [$-9.4$, $+11.7$] & 59.9 & -- \\
\bottomrule
\end{tabular}
\caption{AUROC of holistic \gpt{} and \method{} on factual and judgment labels. The factual label is the \bench{} label. The judgment label is the full \prb{} correctness label on the natural-science reviews and the \rs{} overall factuality rating on the machine learning reviews. $\Delta$ is the paired difference to holistic \gpt{}. \method{} uses the prompt variant of each part (Appendix~\ref{app:prompts}), as in Table~\ref{tab:main}.}
\label{tab:labels}
\end{table*}

\definecolor{stdec}{HTML}{A8702A}\definecolor{stroute}{HTML}{9DBBD6}\definecolor{stver}{HTML}{3E7099}\definecolor{stagg}{HTML}{4A4A4A}\definecolor{stlab}{HTML}{BDBDBD}
\newcommand{\stagebox}[1]{\textcolor{#1}{\rule{0.62em}{0.62em}}}

\begin{figure}[!ht]
\begin{minipage}[c]{0.58\textwidth}
\centering
\includegraphics[width=\linewidth]{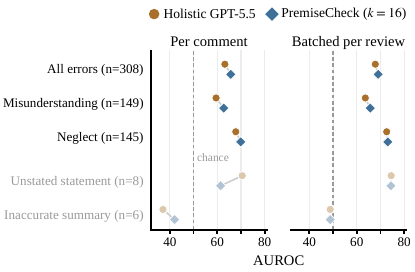}
\end{minipage}\hfill
\begin{minipage}[c]{0.38\textwidth}
\caption{AUROC by error type on the NLP reviews for holistic \gpt{} and \method{} ($k{=}16$), per comment (holistic at medium reasoning) and batched per review (holistic without reasoning). Gray types have too few errors to interpret.}
\label{fig:error-types}
\end{minipage}
\end{figure}

\subsection{Error Types and Scope}
\label{app:scope}

Table~\ref{tab:error-types} and Table~\ref{tab:labels} give the full numbers behind \S\ref{sec:scope}.
Table~\ref{tab:error-types} breaks down AUROC by error type, and Figure~\ref{fig:error-types} plots it.

Table~\ref{tab:labels} compares \method{} with holistic \gpt{} on factual and judgment labels.

\subsection{Failure Analysis}
\label{app:failures}

\begin{figure*}[t]
\centering
\includegraphics[width=\textwidth]{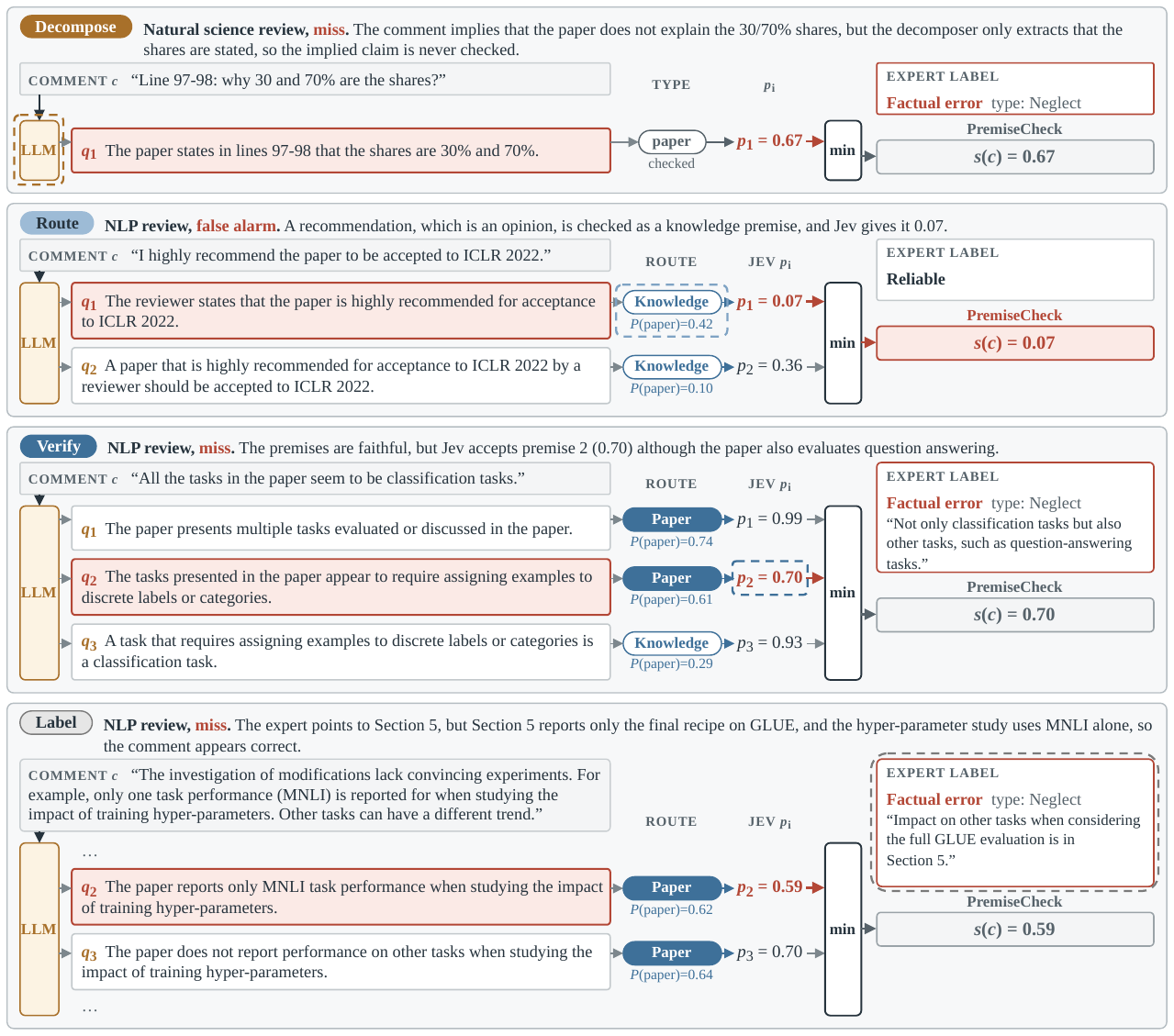}
\caption{Four failures of \method{}, one for each stage. A dashed ring marks the failing component, the lowest premise is framed in red, and the comment, premises, and expert text are verbatim.}
\label{fig:failure-examples}
\end{figure*}

\S\ref{sec:failures} analyzes where \method{} fails.
For each part of \bench{}, we sample false alarms, which are correct comments among the 10\% lowest \method{} scores of that part, and misses, which are factual errors scored above the median, with a fixed seed.
The targets are 50 of each on the NLP reviews, 15 of each on the machine learning reviews, and 20 of each on the natural-science reviews, where only 14 misses qualify and we take all of them.
Each annotator sees the comment, the expert label and rationale, the premises with their types, routes, and probabilities, and the paper text that \jev{} reads, and assigns the earliest stage that explains the failure with one code of Table~\ref{tab:failure-codes}.
Three annotators label all 164 comments independently.
Fleiss' $\kappa$ is 0.45 for the stage and 0.47 for the code.
We take the code chosen by at least two annotators and settle the other 26 comments by discussion.
A comment counts as a failure of holistic \gpt{} under the same rule when \gpt{} scores a correct comment within its own lowest 10\% or a factual error above its own median.
Table~\ref{tab:failure-stages} gives the counts behind Figure~\ref{fig:failures}.
\begin{table}[t]
\begin{minipage}[t]{0.48\textwidth}
\vspace{0pt}
\centering
\footnotesize
\setlength{\tabcolsep}{2.4pt}
\begin{tabular}{@{}llrr@{}}
\toprule
Code & Failure & FA & Miss \\
\midrule
\multicolumn{2}{@{}l}{\stagebox{stdec}\ \textbf{Decompose}} & \textit{40} & \textit{19} \\
\quad D1 & Adds a claim the reviewer did not make & \textbf{21} & \dimrow{0} \\
\quad D2 & Refers to the wrong object & \textbf{12} & \dimrow{0} \\
\quad D3 & Hinges on a vague quantifier & 7 & \dimrow{0} \\
\quad D4 & Erroneous claim not extracted or weakened & \dimrow{0} & \textbf{19} \\
\addlinespace[2pt]
\multicolumn{2}{@{}l}{\stagebox{stroute}\ \textbf{Route}} & \textit{10} & \textit{6} \\
\quad R1 & Opinion or request checked as a fact & 2 & \dimrow{0} \\
\quad R2 & Paper premise routed to knowledge & 3 & 3 \\
\quad R3 & Knowledge premise routed to paper & 5 & \dimrow{0} \\
\quad R4 & Erroneous premise not checked & \dimrow{0} & 3 \\
\addlinespace[2pt]
\multicolumn{2}{@{}l}{\stagebox{stver}\ \textbf{Verify}} & \textit{20} & \textit{27} \\
\quad V1 & A named location misleads \jev{} & 4 & 1 \\
\quad V2 & Evidence lies outside the paper text & \textbf{10} & 8 \\
\quad V3 & Evidence lies after the truncation & \dimrow{0} & \dimrow{0} \\
\quad V4 & \jev{} errs on a clear, faithful premise & 6 & \textbf{18} \\
\addlinespace[2pt]
\multicolumn{2}{@{}l}{\stagebox{stagg}\ \textbf{Aggregate}} & \textit{3} & \textit{2} \\
\quad A1 & A side premise sets the minimum & 3 & \dimrow{0} \\
\quad A2 & Error lies in no single premise & \dimrow{0} & 2 \\
\addlinespace[2pt]
\multicolumn{2}{@{}l}{\stagebox{stlab}\ \textbf{Label}} & \textit{12} & \textit{25} \\
\quad L1 & Expert label questionable & \textbf{10} & \textbf{24} \\
\quad L2 & Undecidable without the full review & 2 & 1 \\
\midrule
 & Total & 85 & 79 \\
\bottomrule
\end{tabular}
\caption{Failure codes and their final counts over all three parts, grouped by stage with the colors of Figure~\ref{fig:failures}. Each code describes the premise, the route, or the check at fault. FA is false alarm, italic numbers are stage totals, and bold marks the three most frequent codes of each case. No comment received the code Other.}
\label{tab:failure-codes}
\end{minipage}\hfill
\begin{minipage}[t]{0.48\textwidth}
\vspace{0pt}
\centering
\footnotesize
\setlength{\tabcolsep}{2.4pt}
\renewcommand{\arraystretch}{1.15}
\begin{adjustbox}{max width=\linewidth}
\begin{tabular}{@{}lcccccc@{}}
\toprule
 & \multicolumn{2}{c}{NLP} & \multicolumn{2}{c}{ML} & \multicolumn{2}{c}{Science} \\
\cmidrule(lr){2-3}\cmidrule(lr){4-5}\cmidrule(l){6-7}
Stage & FA & Miss & FA & Miss & FA & Miss \\
\midrule
\stagebox{stdec}\ Decompose & \cellcolor{stdec!52}\textbf{29}\,{\scriptsize\textbf{(58)}} & \cellcolor{stdec!16}9\,{\scriptsize(18)} & \cellcolor{stdec!30}5\,{\scriptsize(33)} & \cellcolor{stdec!30}5\,{\scriptsize(33)} & \cellcolor{stdec!27}6\,{\scriptsize(30)} & \cellcolor{stdec!32}5\,{\scriptsize(36)} \\
\stagebox{stroute}\ Route & \cellcolor{stroute!13}7\,{\scriptsize(14)} & \cellcolor{stroute!6}2\,{\scriptsize(4)} & \cellcolor{stroute!18}3\,{\scriptsize(20)} & \cellcolor{stroute!6}1\,{\scriptsize(7)} & \dimrow{--} & \cellcolor{stroute!19}3\,{\scriptsize(21)} \\
\stagebox{stver}\ Verify & \cellcolor{stver!13}7\,{\scriptsize(14)} & \cellcolor{stver!32}18\,{\scriptsize(36)} & \cellcolor{stver!24}4\,{\scriptsize(27)} & \cellcolor{stver!24}4\,{\scriptsize(27)} & \cellcolor{stver!40}\textbf{9}\,{\scriptsize\textbf{(45)}} & \cellcolor{stver!32}5\,{\scriptsize(36)} \\
\stagebox{stagg}\ Aggregate & \cellcolor{stagg!6}2\,{\scriptsize(4)} & \cellcolor{stagg!6}1\,{\scriptsize(2)} & \dimrow{--} & \cellcolor{stagg!6}1\,{\scriptsize(7)} & \cellcolor{stagg!6}1\,{\scriptsize(5)} & \dimrow{--} \\
\stagebox{stlab}\ Label & \cellcolor{stlab!9}5\,{\scriptsize(10)} & \cellcolor{stlab!36}\textbf{20}\,{\scriptsize\textbf{(40)}} & \cellcolor{stlab!18}3\,{\scriptsize(20)} & \cellcolor{stlab!24}4\,{\scriptsize(27)} & \cellcolor{stlab!18}4\,{\scriptsize(20)} & \cellcolor{stlab!6}1\,{\scriptsize(7)} \\
\midrule
Sampled & 50 & 50 & 15 & 15 & 20 & 14 \\
\dimrow{\gpt{} also fails} & \dimrow{14} & \dimrow{20} & \dimrow{2} & \dimrow{5} & \dimrow{3} & \dimrow{5} \\
\bottomrule
\end{tabular}
\end{adjustbox}
\caption{Final stage of each sampled failure by part and case, with the share of the cell in percent in parentheses. Shading grows with the share, and bold marks a stage with at least 40\% of a cell. The last row counts comments that holistic \gpt{} also gets wrong under the same rule.}
\label{tab:failure-stages}
\end{minipage}
\end{table}

Figure~\ref{fig:failure-examples} shows one case for each stage, chosen by a fixed rule: among comments whose final code was chosen by at least two annotators, all of them with the highest confidence, we take the shortest comment.
No aggregation failure meets this rule.

\section{Batching and Fusion}
\label{app:batch}

\begin{table*}[tp]
\centering
\footnotesize
\setlength{\tabcolsep}{3.4pt}
\begin{tabular}{@{}lccccc@{}}
\toprule
 & & & & \multicolumn{2}{c}{$\Delta$ [95\% CI]} \\
\cmidrule(l){5-6}
System & AUROC & Cost (\%) & Latency (s) & to reference & to batched holistic (none) \\
\midrule
\multicolumn{6}{@{}l}{\emph{Per comment}} \\
Holistic \gpt{} (medium, ref.) & 63.2 & 100.0 & 6.27 & -- & $-4.6$ [$-6.7$, $-2.3$] \\
\method{} (none, $k{=}16$) & \textbf{65.7} & 24.7 & 2.80 & $+2.4$ [$-0.3$, $+5.3$] & $-2.1$ \\
\rowcolor{ourrow} Fusion: holistic + \method{} (medium) & 64.9 & 151.2 & 6.27 & $+1.6$ [$+0.2$, $+3.0$] & -- \\
\rowcolor{ourrow} Fusion: holistic + \method{} (low) & 65.5 & 134.3 & 6.27 & $+2.3$ [$+0.9$, $+3.5$] & -- \\
\rowcolor{ourrow} Fusion: holistic + \method{} (none) & 64.9 & 124.6 & 6.27 & $+1.7$ [$+0.2$, $+3.2$] & -- \\
\midrule
\multicolumn{6}{@{}l}{\emph{Batched per review}} \\
Holistic \gpt{} (medium) & 64.9 & 19.9 & 1.40 & $+1.7$ [$-0.2$, $+3.5$] & $-2.9$ \\
Holistic \gpt{} (none) & 67.8 & 7.7 & 0.14 & $+4.6$ [$+2.3$, $+6.7$] & -- \\
\rowcolor{ourrow} \method{} (none, $k{=}16$) & 69.0 & 20.2 & 0.92 & $+5.8$ [$+3.1$, $+8.6$] & $+1.2$ [$-1.2$, $+4.0$] \\
\method{} (none, $k{=}1$) & 66.9 & 20.2 & -- & -- & $-0.9$ [$-3.4$, $+1.8$] \\
\method{} ($k{=}16$), \jev{} batched & 68.4 & 14.5 & -- & $+5.2$ & $+0.6$ [$-2.1$, $+3.4$] \\
\rowcolor{ourrow} Fusion: holistic (none) + \method{} & \textbf{71.5} & 28.0 & 0.92 & $+8.3$ & $+3.7$ [$+2.4$, $+5.1$] \\
\midrule
\multicolumn{6}{@{}l}{\emph{Batched \method{} ($k{=}16$) minus}} \\
\quad batched holistic (medium) & \multicolumn{5}{l}{$+4.1$ [$+1.5$, $+6.9$]} \\
\quad per-comment \method{} & \multicolumn{5}{l}{$+3.4$ [$+1.0$, $+5.6$]} \\
\bottomrule
\end{tabular}
\caption{Batched and fused systems on the NLP reviews (6,466 comments, 308 factual errors). Bold marks the best AUROC in each block. Cost is relative to per-comment holistic \gpt{} at medium reasoning. Reasoning in parentheses refers to the holistic call or to the decomposer. Per-comment fusion combines holistic \gpt{} (medium) with \method{} using a single question per premise. Fusion runs both systems in parallel.}
\label{tab:batch}
\end{table*}

Table~\ref{tab:batch} compares per-comment and batched systems and their fusion.
Batched holistic judgment uses the prompt and question of holistic \gpt{}, with all comments of one review numbered in the evidence and one question per comment.
Batched \method{} uses the decomposition prompt of Appendix~\ref{app:prompts}, changed only so that it returns one premise list per numbered comment.
Both batched systems see the same comments of each review.
When an answer misses comments, we ask once more for the missing ones, after which no comment remained missing.
The cost of one call is divided evenly among the comments of the review.
In the row ``\jev{} batched'', \jev{} also reads the paper once for up to eight premises, with two questions per premise.
Latency is per comment, computed as the time for the whole review divided by its number of comments.

\section{Examples}
\label{app:examples}

\subsection{Scored Comments}
\label{app:scored}

\begin{figure*}[tp]
\centering
\includegraphics[width=\textwidth]{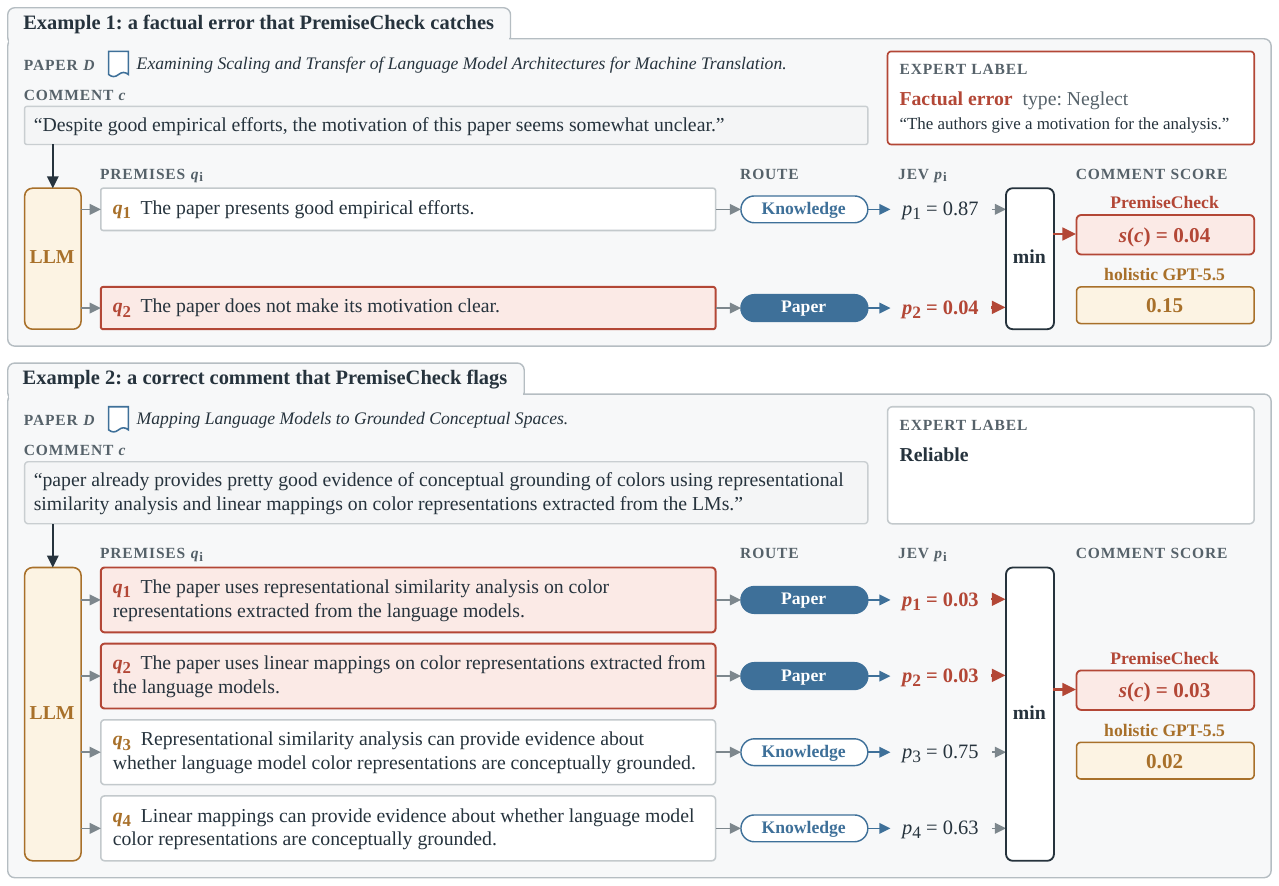}
\caption{Two comments from the NLP reviews scored by \method{} with the medium-reasoning decomposer. Route and $p$ are the decision and probability of \jev{}, red marks the premise with the lowest probability, which gives the comment score, and the comment, premises, and expert text are verbatim.}
\label{fig:examples}
\end{figure*}

Figure~\ref{fig:examples} shows two comments from the NLP reviews with the premises of \method{} (medium-reasoning decomposer) and the probabilities that \jev{} assigns to them.

The first comment is a factual error.
We drew it at random from short factual errors in human-written reviews that both systems flag, that is, errors in the lowest 10\% of \method{} scores that holistic \gpt{} also scores at or below its median.
The reviewer claims that the motivation of the paper is unclear.
The expert disagrees, because the authors state a motivation for their analysis.
\method{} splits the comment into two premises.
\jev{} checks the second premise, that the paper does not make its motivation clear, against the paper and gives it a probability of 0.04.
This premise states the point that the expert disputes, and it sets the comment score.
Holistic \gpt{} also flags the comment, with a probability of 0.15.

The second comment is correct, but \method{} flags it.
We drew it at random from short correct comments of human-written reviews in the lowest 10\% of \method{} scores.
The comment describes evidence from representational similarity analysis and linear mappings, but the paper text contains neither.
The comment therefore likely refers to another paper.
The decomposer rewrote ``paper'' as ``the paper'', so premises 1 and 2 are false as written, and \jev{} gives both a probability of 0.03.
Holistic \gpt{} also gives the comment a very low score.

\begin{figure*}[t]
\centering
\includegraphics[width=\textwidth]{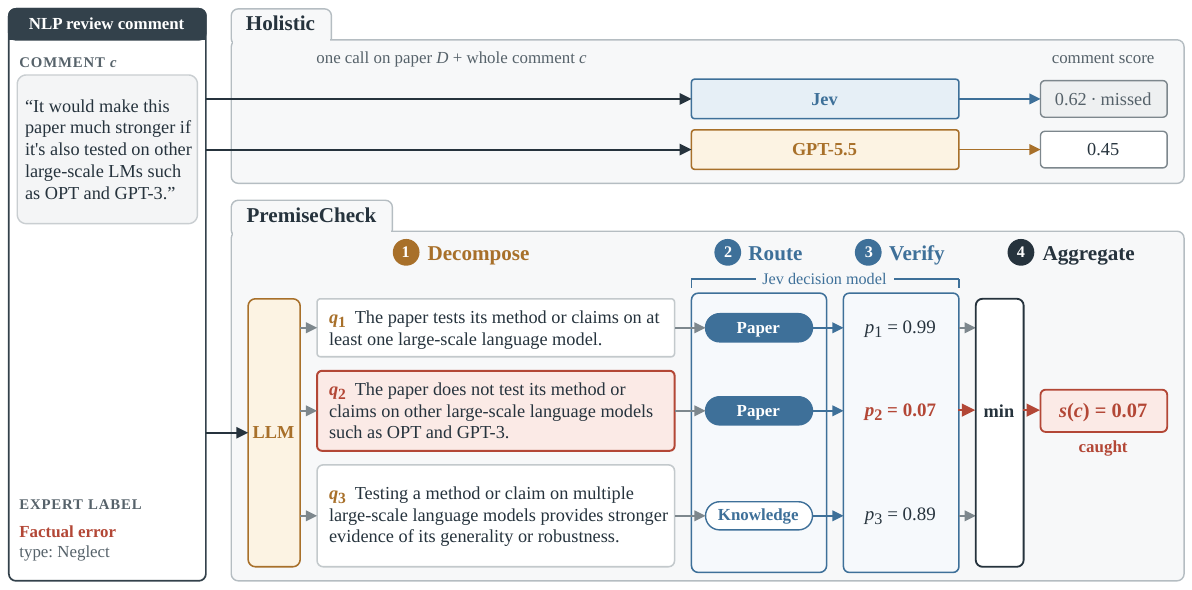}
\caption{A factual error from the NLP reviews, chosen as an illustration because its holistic and decomposed scores differ strongly. Holistic \jev{} gives the comment 0.62, whereas \method{} with the no-reasoning decomposer gives premise 2 a probability of 0.07, which sets the comment score. Comment and premises are verbatim.}
\label{fig:decomposition-example}
\end{figure*}

\begin{figure*}[t]
\centering
\includegraphics[width=\textwidth]{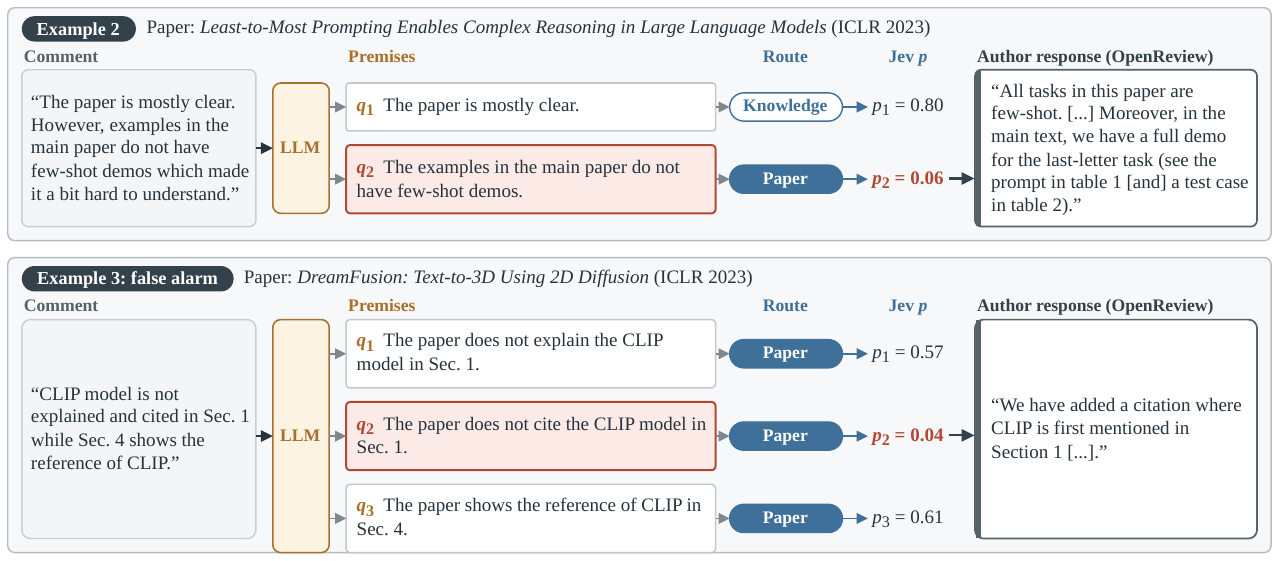}
\caption{Two more comments from the machine learning reviews flagged by \method{}, with the authors' OpenReview responses. Route and $p$ are the decision and probability of \jev{}, red marks the premise with the lowest probability, and quotes are verbatim except for omissions marked [\ldots] and a typo corrected in brackets.}
\label{fig:rebuttal}
\end{figure*}

Figure~\ref{fig:decomposition-example} shows a further factual error, for which \method{} gives a much lower score than holistic \jev{}.
We chose it among NLP factual errors with a \method{} score at or below 0.10 and a holistic \jev{} score at or above 0.60.
The comment implies that the paper does not test other large-scale language models such as OPT and GPT-3.

\subsection{Flagged Comments with Author Responses}
\label{app:rebuttal}

We choose the examples by a fixed rule among the 10\% of machine learning comments that \method{} finds most suspicious.
The two caught errors are the shortest comments that the \rs{} annotators label as misinformed and whose authors reply that the paper contains what the comment asks for.
Example~1 is Figure~\ref{fig:flag-example}, and Example~2 in Figure~\ref{fig:rebuttal} is the second.
The false alarm is the shortest comment that the annotators label as correct and whose authors accept it.
In Example~3, the reviewer says that Section~1 does not cite CLIP, and the authors confirm this by adding the citation.
\jev{} still gives the premise a probability of 0.04, probably because the paper cites CLIP in Section~4 and \jev{} reads the whole paper, so it cannot tell in which section a citation appears.

\end{document}